\documentclass[10pt,journal,compsoc]{IEEEtran}
\usepackage[
            pdfstartview=FitH,
            CJKbookmarks=true,
            bookmarksnumbered=true,
            bookmarksopen=true,
            colorlinks, 
            pdfborder=001,   
            linkcolor=green,
            anchorcolor=green,
            citecolor=green
            ]{hyperref}
  \usepackage[nocompress]{cite}

\usepackage{graphicx}
\usepackage{bbm}
\usepackage{amsmath,amssymb} 
\usepackage{caption}
\usepackage{color}
\usepackage{cite}
\usepackage{bm}
\usepackage{microtype}
\usepackage[tight,footnotesize]{subfigure}
\usepackage{multirow}
\usepackage{floatrow}
\usepackage{colortbl}
\usepackage{array}
\definecolor{mygray}{gray}{.9}

\ifCLASSINFOpdf
\usepackage{bm}
\else
\fi

\begin{document}

\title{Person Re-identification: \\Past, Present and Future}

\author{Liang Zheng,
        Yi Yang, 
        and Alexander G. Hauptmann

\IEEEcompsocitemizethanks{
\IEEEcompsocthanksitem L. Zheng and Y. Yang are with the
Centre for Quantum Computation and Intelligent Systems, University of Technology at Sydney, NSW, Australia. \protect\\
E-mail: liangzheng06@gmail.com, yee.i.yang@gmail.com
\IEEEcompsocthanksitem A. Hauphtmann is with the School of Computer Science at Carnegie Mellon University, with a joint appointment in the Language Technologies Institute.\protect\\
E-mail: alex@cs.cmu.edu
}
}

\markboth{Journal of \LaTeX\ Class Files,~Vol.~14, No.~8, August~2015}%
{SYan2016hell \MakeLowercase{\textit{et al.}}: Bare Advanced Demo of IEEEtran.cls for IEEE Computer Society Journals}

\IEEEtitleabstractindextext{%
\begin{abstract}
Person re-identification (re-ID) has become increasingly popular in the community due to its application and research significance. It aims at spotting a person of interest in other cameras. In the early days, hand-crafted algorithms and small-scale evaluation were predominantly reported. Recent years have witnessed the emergence of large-scale datasets and deep learning systems which make use of large data volumes. Considering different tasks, we classify most current re-ID methods into two classes, \emph{i.e.,} image-based and video-based; in both tasks, hand-crafted and deep learning systems will be reviewed. Moreover, two new re-ID tasks which are much closer to real-world applications are described and discussed, \emph{i.e.,} end-to-end re-ID and fast re-ID in very large galleries. This paper: 1) introduces the history of person re-ID and its relationship with image classification and instance retrieval; 2) surveys a broad selection of the hand-crafted systems and the large-scale methods in both image- and video-based re-ID; 3) describes critical future directions in end-to-end re-ID and fast retrieval in large galleries; and 4) finally briefs some important yet under-developed issues.

\end{abstract}

\begin{IEEEkeywords}
Large-scale person re-identification, hand-crafted systems, Convolutional Neural Network, literature survey.
\end{IEEEkeywords}}

\maketitle

\IEEEdisplaynontitleabstractindextext

\IEEEpeerreviewmaketitle

\ifCLASSOPTIONcompsoc
\IEEEraisesectionheading{\section{Introduction}\label{sec:introduction}}
\else
\section{Introduction}
\label{sec:introduction}
\fi

\IEEEPARstart{A}{ccording} to Homer (\emph{Odyssey} \romannumeral4:412), \emph{Mennelaus} was becalmed on his journey home from the Trojan War; He wanted to propitiate the gods and return safely home. He was told that he should capture \emph{Proteus} and force him to reveal the answer. Although \emph{Proteus} transformed to a lion, a serpent, a leopard, water and also a tree, \emph{Mennelaus} then succeeded in holding him as he emerged from the sea to sleep among the seals. \emph{Proteus} was finally compelled to answer to him truthfully.

Perhaps this is one of the oldest stories about re-identifying a person even after intensive appearance changes. In 1961, when discussing the relationship between mental states and behavior, Alvin Plantinga \cite{plantinga1961things} provided one of the first definitions of re-identification:
\begin{quote}
``To re-identify a particular, then, is to identify it as (numerically) the same particular as one encountered on a previous occasion''.
\end{quote}
Person re-identification had thus been studied in various research and documentation areas such as metaphysics \cite{plantinga1961things}, psychology \cite{rorty1973transformations}, and logic \cite{cocchiarella1977sortals}. All these works are grounded on Leibniz's Law which claims that ``there cannot be separate objects or entities that have all their properties in common.''

In the modern computer vision community, the task of person re-ID shares similar insights with the old times. In video surveillance, when being presented with a person-of-interest (query), person re-ID tells whether this person has been observed in another place (time) by another camera. The emergence of this task can be attributed to 1) the increasing demand of public safety and 2) the widespread large camera networks in theme parks, university campuses and streets, \emph{etc}. Both causes make it extremely expensive to rely solely on brute-force human labor to accurately and efficiently spot a person-of-interest or to track a person across cameras.

\begin{figure}[t]
  \centering
  \includegraphics[width=3.2in]{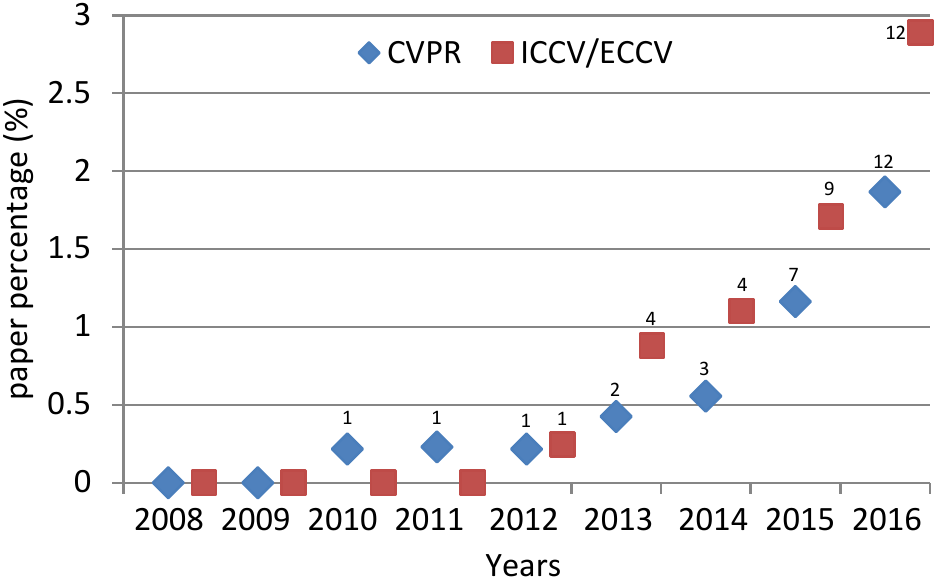}\\
  \caption{Percentage of person re-ID papers on top conferences over the years. Numbers above the markers indicate the number of re-ID papers. }\label{fig:top-conf}
\end{figure}

Technically speaking, a practical person re-ID system in video surveillance can be broken down into three modules, \emph{i.e.,} person detection, person tracking, and person retrieval. It is generally believed that the first two modules are independent computer vision tasks, so most re-ID works focus on the last module, \emph{i.e.,} person retrieval. In this survey, if not specified, person re-ID refers to the person retrieval module. From the perspective of computer vision, the most challenging problem in re-ID is how to correctly match two images of the same person under intensive appearance changes, such as lighting, pose, and viewpoint, which has important scientific values. Given its research and application significance, the re-ID community is fast growing, evidenced by an increasing number of publications in the top venues (Fig. \ref{fig:top-conf}).

\subsection{Organization of This Survey}
Some person re-ID surveys exist \cite{d2012people,bedagkar2014survey,gong2014person,satta2013appearance}. In this survey, we mainly discuss the vision part of re-ID, which is also a focus in the community, and refer readers to the camera calibration and view topology methods in \cite{bedagkar2014survey}. Another difference from previous surveys is that we focus on different re-ID subtasks currently available or likely to be visible in the future, instead of very detailed techniques or architectures. Special emphasis is given deep learning methods, end-to-end re-ID and very large scale re-ID, which are currently popular topics or reflect future trends. This survey first introduces a brief history of person re-ID in Section \ref{sec:history} and its relationship with classification and retrieval in Section \ref{sec:relationship}.
We then describe previous literature in image-based and video-based person re-ID in Section \ref{sec:image_reid} and Section \ref{sec:video_reid}, respectively. Both sections categorize methods into hand-crafted and deeply-learned systems.  In Section \ref{sec:det_track_reid}, since the relationship between detection, tracking, and re-ID has not been extensively studied, we will discuss several previous works and point out future research emphasis. In Section \ref{sec:scale_reid}, large-scale re-ID which resorts to state-of-the-art retrieval models will be introduced, which is also an important future direction. Some other open issues will be summarized in Section \ref{sec:open}, and conclusions will be drawn in Section \ref{sec:conclusion}.
\begin{figure*}[t]
  \centering
  \includegraphics[width=7.1in]{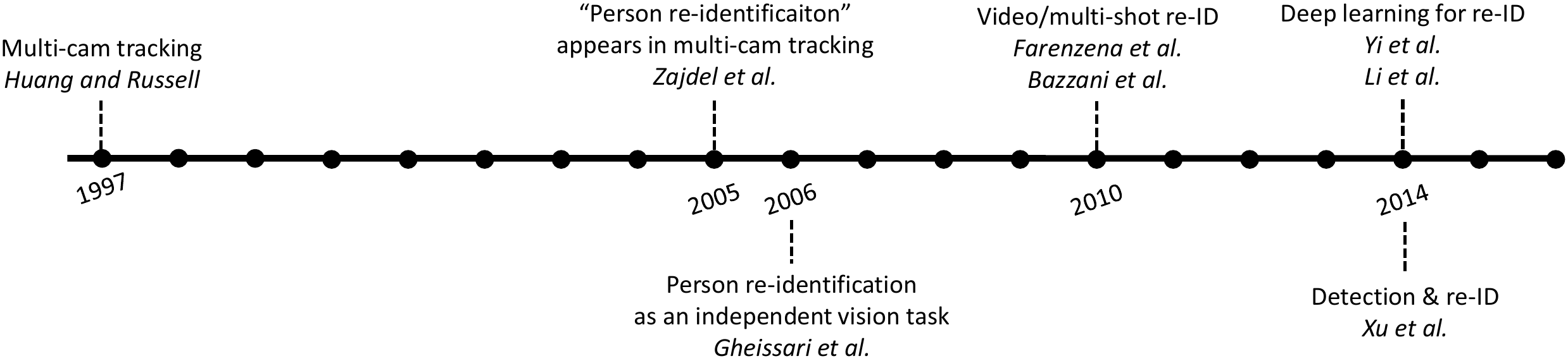}\\
  \caption{Milestones in the person re-ID history. }\label{fig:history}
\end{figure*}

\subsection{A Brief History of Person Re-ID}\label{sec:history}
Person re-ID research started with multi-camera tracking \cite{wang2013intelligent}. Several important re-ID directions have been developed since then. In this survey, we briefly introduce some milestones in person re-ID history (Fig. \ref{fig:history}).

\textbf{Multi-camera tracking.} In the early years, person re-ID, as a the term without being formally raised, was tightly twined with multi-camera tracking, in which appearance models were integrated with the geometry calibration among disjoint cameras. In 1997, Huang and Russell \cite{huang1997object} proposed a Bayesian formulation to estimate the posterior of predicting the appearance of objects in one camera given evidence observed in other camera views. The appearance model includes multiple spatial-temporal features such as color, vehicle length, height and width, velocity, and time of observation. A comprehensive survey of multi-camera tracking can be accessed in \cite{wang2013intelligent}.

\textbf{Multi-camera tracking with explicit ``re-identification''.} To our knowledge, the first work on multi-camera tracking where the term ``person re-identification'' is proposed, was published in 2005 by Wojciech Zajdel, Zoran Zivkovic and Ben J. A. Kr\"{o}se from the University of Amsterdam \cite{zajdel2005keeping}. In their ICRA'05 paper entitled ``Keeping track of humans: Have I seen this person before?'', Zajedel \emph{et al.} aims to ``re-identify a person when it leaves the field of view and re-enters later''. In their method, a unique, latent label is assumed for every person, and a dynamic Bayesian network is defined to encode the probabilistic relationship between the labels and features (color and spatial-temporal cues) from the tracklets. The ID of an incoming person is determined by the posterior label distributions computed by an approximate Bayesian inference algorithm.

\textbf{The independence of re-ID (image-based).} One year later in 2006, Gheissari \emph{et al.} \cite{gheissari2006person} employed only the visual cues of persons after a spatial-temporal segmentation algorithm for foreground detection. Visual matching based on color and salient edgel histograms is performed by either an articulated pedestrian model or the Hessian-Affine interest point operator. Experiments are conducted on a dataset with 44 persons captured by 3 cameras with moderate view overlap. Note that, although Gheissari \emph{et al.} \cite{gheissari2006person} design a spatial-temporal segmentation method using the video frames, neither the feature design nor matching processes use the video information, so we classify \cite{gheissari2006person} into image-based re-ID. This work \cite{gheissari2006person} marks the separation of person re-ID from multi-camera tracking, and its beginning as an independent computer vision task.

\textbf{Video-based re-ID.} Initially intended for tracking in videos, most re-ID works focus on image matching instead. In the year 2010, two works \cite{bazzani2010multiple,farenzena2010person} were proposed for multi-shot re-ID, in which frames are randomly selected. Color is a common feature used in both works, and Farenzena \emph{et al.} \cite{farenzena2010person} additionally employ a segmentation model to detect the foreground. For distance measurement, both works calculate the minimum distance among bounding boxes in two image sets, and Bazzani \emph{et al.} further use the Bhattacharyya distance for the color and generic epitome features. It is shown that using multiple frames per person effectively improves over the single-frame version \cite{bazzani2010multiple,farenzena2010person} and that re-ID accuracy will saturate as the number of selected frames increases \cite{bazzani2010multiple}.

\textbf{Deep learning for re-ID.} The success of deep learning in image classification \cite{krizhevsky2012imagenet} spreads to re-ID in 2014, when Yi \emph{et al.} \cite{yi2014deep} and Li \emph{et al.} \cite{li2014deepreid} both employ a siamese neural network \cite{bromley1993signature} to determine if a pair of input images belong to the same ID. The reason for choosing the siamese model is probably that the number of training samples for each identity is limited (usually two). Aside from some variations in parameter settings, the main differences are that \cite{yi2014deep} adds an additional cost function in the network, while \cite{li2014deepreid} uses a finer body partitioning. The experimental datasets do not overlap in \cite{yi2014deep} and \cite{li2014deepreid}, so the two methods are not directly comparable. Although its performance is not stable yet on the small datasets, deep learning methods has since become a popular option in re-ID

\textbf{End-to-end image-based re-ID.} While a majority of works use hand-cropped boxes or boxes produced by a fixed detector in their experiments, it is necessary to study the impact of pedestrian detectors on re-ID accuracy. In 2014, Xu \emph{et al.} \cite{xu2014person} addressed this topic by combining the detection (commonness) and re-ID (uniqueness) scores. It is shown that on the CAMPUS dataset, jointly considering detection and re-ID confidence leads to higher person retrieval accuracy than using them separately.

\subsection{Relationship with Classification and Retrieval} \label{sec:relationship}
Person re-ID lies inbetween image classification \cite{krizhevsky2012imagenet} and instance retrieval \cite{zheng2016sift} in terms of the relationship between training and testing classes (Table \ref{table:compare_task}). For image classification, training images are available for each class, and testing images fall into these predefined classes, denoted as previously ``seen'' in Table \ref{table:compare_task}. For instance retrieval, usually there is no training data because one does not know  the content of the query in advance and the gallery may contain various types of objects. So the training classes are ``not available'' and the testing classes (queries) are denoted as previously ``unseen''.

Compared to image classification, person re-ID is similar in that the training classes are available, which includes images of different identities. Person re-ID is also similar to instance retrieval in that the testing identities are unseen: they do not have overlap with the training identities, except that both training and testing images are of pedestrians.

As a consequence, person re-ID can be positioned to take advantage of both classification and retrieval. On the one hand, using training classes, discriminative distance metrics \cite{liao2015person} or feature embeddings \cite{li2014deepreid,zheng2016mars} can be learned in the person space. On the other hand, when it comes to retrieval, efficient indexing structures \cite{AKM} and hashing techniques \cite{wang2016survey} can be beneficial for re-ID in a large gallery. In this survey, both effective learning  and efficient retrieval approaches will be introduced or pointed out as important future directions.

\setlength{\tabcolsep}{3.55pt}
\begin{table}[t]
\centering
\small
\caption{Comparing re-ID with classification and retrieval}
\begin{tabular}{l|lll}
\hline
Task&  Train Class & Test Class & Advantage\\
\hline
Classification& available & seen & discri. learning \\
Retrieval& not available & unseen & efficiency \\
Person re-ID&  available & unseen &  discri. + efficiency?\\
\hline
\end{tabular}
\label{table:compare_task}
\end{table}

\section{Image-based Person Re-ID} \label{sec:image_reid}
Since the work by Gheissari \emph{et al.}  in 2006 \cite{gheissari2006person}, person re-ID has mostly been explored using single images. Let us consider a closed-world toy model, in which $\mathcal{G}$ is a gallery (database) composed of $N$ images, denoted as $\{g_i\}_{i=1}^N$. They belong to $N$ different identities $1, 2, ..., N$. Given a probe (query) image $q$, its identity is determined by:
\begin{equation}\label{eq:img_reid}
i^* = {\arg\max}_{i\in 1,2,...,N} \mbox{sim} (q, g_i),
\end{equation}
where $i^*$ is the identity of probe $q$, and $\mbox{sim}(\cdot,\cdot)$ is some kind of similarity function. 

\subsection{Hand-crafted Systems}
It is apparent from Eq. \ref{eq:img_reid} that two components are necessary for a toy re-ID system, \emph{i.e.,} image description and distance metrics.

\subsubsection{Pedestrian Description}
In pedestrian descriptions, the most commonly used feature is color, while texture features are less frequent. In \cite{farenzena2010person}, the pedestrian foreground is segmented from the background, and a symmetrical axis is computed for each body part. Based on body configuration, the weighted color histogram (WH), the maximally stable color regions (MSCR), and the recurrent high-structured patches (RHSP) are computed. WH assigns larger weights to pixels near the symmetrical axis and forms a color histogram for each part. MSCR detects stable color regions and extracts features such as color, area, and centroid. RHSP instead is a texture feature capturing recurrent texture patches. Gheissari \emph{et al.} \cite{gheissari2006person} propose a spatial-temporal segmentation method to detect stable foreground regions. For a local region, an HS histogram and an edgel histogram are computed. The latter encodes the dominant local boundary orientation and the RGB ratios on either sides of the edgel. Gray and Tao \cite{gray2008viewpoint} use 8 color channels (RGB, HS, and YCbCr) and 21 texture filters on the luminance channel, and the pedestrian is partitioned into horizontal stripes. A number of later works \cite{prosser2010person,zheng2013reidentification,ma2013domain} employ the same set of features as \cite{gray2008viewpoint}. Similarly, Mignon \emph{et al.} \cite{mignon2012pcca} build the feature vector from RGB, YUV and HSV channels and the LBP texture histograms in horizontal stripes.

Compared to the earlier works described above, hand-crafted features have remained more or less the same in recent years \cite{zheng2015partial,zhao2013unsupervised,li2013learning,chen2016similarity,liao2015person}. In a series of works by Zhao \emph{et al.} \cite{zhao2013unsupervised,zhao2013person,zhao2014learning}, the 32-dim LAB color histogram and the 128-dim SIFT descriptor are extracted from each $10\times 10$ patch densely sampled with a step size of 5 pixels; this feature is also used in \cite{shen2015person}. Adjacency constrained search is employed to find the best match for a query patch in horizontal stripes with similar latitudes in a gallery image. Das \emph{et al.} \cite{das2014consistent} apply HSV histograms on the head, torso and legs from the silhouette proposed in \cite{bazzani2010multiple}.  Li \emph{et al.} \cite{li2013learning} also extract local color descriptors from patches but aggregate them using hierarchical Gaussianization \cite{zhou2009hierarchical} to capture spatial information, a procedure followed by \cite{chen2015similarity}. Pedagadi \emph{et al.} \cite{pedagadi2013local} extract color histograms and moments from HSV and YUV spaces before dimension reduction using PCA. Liu \emph{et al.} \cite{liu2014semi} extract the HSV histogram, gradient histogram and the LBP histogram for each local patch. To improve the robustness of the RGB values against photometric variance, Yang \emph{et al.} \cite{yang2014salient} introduce the salient color names based color descriptor (SCNCD) for global pedestrian color descriptions. The influence of the background and different color spaces are also analysed. In \cite{liao2015person}, Liao \emph{et al.} propose the local maximal occurrence (LOMO) descriptor, which includes the color and SILTP histograms. Bins in the same horizontal stripe undergo max pooling and a three-scale pyramid model is built before a log transformation. LOMO is later employed by \cite{zhang2016learning,zhangsample} and a similar set of features is used by Chen \emph{et al.} \cite{chen2016similarity}.  In \cite{zheng2015scalable}, Zheng \emph{et al.} propose extracting the 11-dim color names descriptor \cite{van2009learning} for each local patch, and aggregating them into a global vector through a Bag-of-Words (BoW) model. In \cite{matsukawa2016hierarchical}, a hierarchical Gaussian feature is proposed to describe color and texture cues, which models each region by multiple Gaussian distributions. Each distribution represents a patch inside the region.


Apart from directly using low-level color and texture features, another good choice is the attribute-based features which can be viewed as mid-level representations. It is believed that attributes are more robust to image translations compared to low-level descriptors. In \cite{layne2012person}, Layne \emph{et al.} annotate 15 binary attributes on the VIPeR dataset related to attire and soft biometrics. The low-level color and texture features are used to train the attribute classifiers. After attribute weighting, the resulting vector is integrated in the SDALF \cite{farenzena2010person} framework to fuse with other visual features. Liu \emph{et al.} \cite{liu2012attribute} improve the latent Dirichlet allocation (LDA) model using annotated attributes to filter out noisy LDA topics. Liu \emph{et al.} \cite{liu2012person} propose discovering some pedestrian prototypes with common attributes in an unsupervised manner and adaptively determine the feature weights of different query person according to the prototypes. Some recent works borrow external data for attribute learning. In \cite{su2015multi}, Su \emph{et al.} embed the binary semantic attributes of the same person but different cameras into a continuous low-rank attribute space, so that the attribute vector is more discriminative for matching. Shi \emph{et al.} \cite{shi2015transferring} propose learning a number of attributes including color, texture, and category labels from existing fashion photography datasets. These attributes are directly transferred to re-ID under surveillance videos and achieve competitive results. Recently, Li \emph{et al.} \cite{li2016richly} collected a large-scale dataset with richly annotated pedestrian attributes to facilitate attribute-based re-ID methods.

\subsubsection{Distance Metric Learning}
In hand-crafted re-ID systems, a good distance metric is critical for its success, because the high-dimensional visual features typically do not capture the invariant factors under sample variances.  A comprehensive survey of the metric learning methods can be accessed in \cite{yang2006distance}. These metric learning methods are categorized w.r.t supervised learning versus unsupervised learning, global learning versus local learning, \emph{etc}. In person re-ID, the majority of works fall into the scope of supervised global distance metric learning.

The general idea of global metric learning is to keep all the vectors of the same class closer while pushing vectors of different classes further apart. The most commonly used formulation is based on the class of Mahalanobis distance functions, which generalizes Euclidean distance using linear scalings and rotations of the feature space. The squared distance between two vectors $x_i$ and $x_j$ can be written as,
\begin{equation}\label{eq:mahal}
  d(x_i, x_j) = (x_i-x_j)^\mathrm{ T } \bm{\mbox{M}} (x_i-x_j),
\end{equation}
where $\bm{\mbox{M}}$ is a positive semidefinite matrix. Equation \ref{eq:mahal} can be formulated into the convex programming problem suggested by Xing \emph{et al.} \cite{xing2003distance}.

In person re-ID, currently the most popular metric learning method, \emph{i.e.,} KISSME \cite{kostinger2012large} is based on Eq. \ref{eq:mahal}. In this method \cite{kostinger2012large}, the decision on whether a pair $(i, j)$ is similar or not is formulated as a likelihood ratio test. The pairwise difference ($x_{i,j} = x_i-x_j$) is employed and the difference space is assumed to be a Gaussian distribution with a zero mean. It is shown in \cite{kostinger2012large} that the Mahalanobis distance metric can be naturally derived from the log-likelihood ratio test and in practice, the principle component analysis (PCA) is applied to the data points to eliminate dimension correlations.

Based on Eq. \ref{eq:mahal}, a number of other metric learning methods have been introduced. In the early days, some classic metric learning methods target at nearest neighbor classification. Weinberger \emph{et al.} \cite{weinberger2005distance} propose the large margin nearest neighbor
Learning (LMNN) method which sets up a perimeter for the target neighbors (matched pairs) and punishes those invading the perimeter (imposters). This method belongs to the supervised local distance metric learning category \cite{yang2006distance}. To avoid the overfitting problems encountered in LMNN, Davis \emph{et al.} \cite{davis2007information} propose the information-theoretic metric learning (ITML) as a trade-off between satisfying the given similarity constraints and ensuring that the learned metric is close to the initial distance function.

In recent years, Hirzer \emph{et al.} \cite{hirzer2012relaxed} proposed relaxing the positivity constraint which provides a sufficient approximation for the matrix $\bm{\mbox{M}}$ with a much lower computational cost. Chen \emph{et al.} \cite{chen2015similarity} add a bilinear similarity in addition to the Mahalanobis distance, so that cross-patch similarities can be modeled. In \cite{li2013learning}, the global distance metric is coupled with the local adaptive threshold rule which additionally contains the orthogonal information of $(x_i, x_j)$. In \cite{liao2015efficient}, Liao \emph{et al.} suggest perserving with a positive semidefinite constraint and propose weighting the positive and negative samples differently. Yang \emph{et al.} \cite{yang2016large} consider both the differences and commonness between image pairs and show that the covariance matrices of dissimilar pairs can be inferred from those of the similar pairs, which makes the learning process scalable to large datasets.

Other than learning distance metrics, some works focus on learning discriminative subspaces.  Liao \emph{et al.} \cite{liao2015person} propose learning the projection $\bm w$ to a low-dimensional subspace with cross-view data solved in a similar manner to linear discriminant analysis (LDA) \cite{scholkopft1999fisher},
\begin{equation}\label{eq:lda}
  \mathcal{J}(\bm w) = \frac{\bm w^ \mathrm{ T } \bm S_b \bm w}{\bm w^ \mathrm{ T } \bm S_w \bm w},
\end{equation}
where $\bm S_b$ and $\bm S_w$ are the between-class and within-class scatter matrices, respectively. Then, a distance function is learned in the resulting subspace using KISSME. To learn $\bm w$, Zhang \emph{et al.} \cite{zhang2016learning} further employ the null Foley-Sammon transform to learn a discriminative null space which satisfies a zero within-class scatter and a positive between-class scatter. For dimension reduction, Pedagadi \emph{et al.} \cite{pedagadi2013local} sequentially combine the unsupervised PCA (principle component analysis) and supervised local Fisher discriminative analysis which preserves the local neighborhood structure. In \cite{mignon2012pcca}, the pairwise constrained component
analysis (PCCA) is proposed which learns a linear mapping function to be able to work directly on high-dimensional data, while ITML and KISSME should be preceded by a step of dimension reduction.   In \cite{xiong2014person}, Xiong \emph{et al.} further propose improved versions of two existing subspace projection methods, \emph{i.e.,} regularized PCCA \cite{mignon2012pcca} and kernel LFDA \cite{pedagadi2013local}.

Aside from the methods that use Mahalanobis distance (Eq. \ref{eq:mahal}), some use other learning tools such as support vector machine (SVM) or boosting. Prosser \emph{et al.} \cite{prosser2010person} propose learning a set of weak RankSVMs which are subsequently assembled into a stronger ranker. In \cite{liu2015ensemble}, a structural SVM is employed to combine different color descriptors at decision level. In \cite{zhangsample}, Zhang \emph{et al.} learn a specific SVM for each training identity and map each testing image to a weight vector inferred from its visual features. Gray and Tao \cite{gray2008viewpoint} propose using the AdaBoost algorithm to select and combine many different kinds of simple features into a single similarity function.

\subsection{Deeply-learned Systems}
CNN-based deep learning models have been popular since Krizhevsky \emph{et al.} \cite{krizhevsky2012imagenet} won ILSVRC'12 by a large margin. The first two works in re-ID to use deep learning were \cite{yi2014deep,li2014deepreid} as mentioned in Section \ref{sec:history} and Fig. \ref{fig:history}. Generally speaking, two types of CNN models are commonly employed in the community. The first type is the classification model as used in image classification \cite{krizhevsky2012imagenet} and object detection \cite{girshick2013rich}. The second is the siamese model using image pairs \cite{radenovic2016cnn} or triplets \cite{schroff2015facenet} as input.
\setlength{\tabcolsep}{1.45pt}
\begin{table}[t]
\centering
\caption{Comparison of the identification and verification (siamese) models on the Market-1501 dataset (single query). }
\begin{tabular}{l|cc|cc}
\hline
Model&  \multicolumn{2}{c|}{\emph{Identification}} & \multicolumn{2}{c}{\emph{Verification}} \\
&  rank-1 (\%)& mAP (\%) &rank-1 (\%)& mAP (\%)\\
\hline
AlexNet \cite{krizhevsky2012imagenet}& 56.03 & 32.38& 41.24 & 22.47 \\
VGG-16 \cite{simonyan2014very}& 64.34  & 40.77 &42.99 &24.29 \\
Residual-50 \cite{he2016deep}& 72.54  & 46.00&60.12 &40.54  \\
\hline
\end{tabular}
\label{table:identification_verification}
\end{table}
The major bottleneck of deep learning in re-ID is the lack of training data. Most re-ID datasets provide only two images for each identity such as VIPeR \cite{gray2008viewpoint}, so currently most CNN-based re-ID methods focus on the siamese model. In \cite{yi2014deep}, an input image is partitioned into three overlapping horizontal parts, and the parts go through two convolutional layers plus a fully connected layer which fuses them and outputs a vector for this image. The similarity of the two output vectors are computed using the cosine distance. The architecture designed by Li \emph{et al.} \cite{li2014deepreid} is different in that a patch matching layer is added which multiplies the convolution responses of two images in different horizontal stripes, similar to ACS \cite{zhao2013unsupervised} in spirit. Later, Ahmed \emph{et al.} \cite{ahmed2015improved} improved the siamese model by computing the cross-input neighborhood difference features, which compares the features from one input image to features in neighboring locations of the other image. While \cite{li2014deepreid} uses product to compute patch similarity in similar latitude, Ahmed \emph{et al.} \cite{ahmed2015improved} use subtraction. Wu \emph{et al.} \cite{wu2016personnet} deepen the networks using convolutional filters of smaller sizes, called ``PersonNet''. In \cite{varior2016siamese}, Varior \emph{et al.} incorporate long short-term memory (LSTM) modules into a siamese network. LSTMs process image parts sequentially so that the spatial connections can be memorized to enhance the discriminative ability of the deep features. Varior \emph{et al.} \cite{varior2016gated} propose inserting a gating function after each convolutional layer to capture effective subtle patterns when a pair of testing images are fed into the network. This method achieves state-of-the-art accuracy on several benchmarks, but its disadvantage is also obvious. The query has to pair with each gallery image before being sent into the network - a time inefficient process in large datasets. Similar to \cite{varior2016gated}, Liu \emph{et al.} \cite{liu2016end} propose integrating a soft attention based model in a siamese network to adaptively focus on the important local parts of an input image pair; however, this method is also limited by computational inefficiency. While these works use image pairs, Cheng \emph{et al.} \cite{cheng2016person} design a triplet loss function that takes three images as input. After the first convolutional layer, four overlapping body parts are partitioned for each image and fused with a global one in the FC layer. Su \emph{et al.} \cite{su2016deep} propose a three-stage learning process which includes attribute prediction using an independent dataset and an attributes triplet loss trained on datasets with ID labels.

A drawback of the siamese model is that it does not make full use of re-ID annotations. In fact, the siamese model only needs to consider pairwise (or triplet) labels. Telling whether an image pair is similar (belong to the same identity) or not is a weak label in re-ID. Another potentially effective strategy consists of using a classification/identification mode, which makes full use of the re-ID labels. In \cite{xiao2016learning}, training identities from multiple datasets jointly form the training set and a softmax loss is employed in the classification network. Together with the proposed impact score for each FC neuron and a domain guided dropout based on the impact score, the learned generic embeddings yield competitive re-id accuracy. On larger datasets, such as PRW and MARS, the classification model achieves good performance without careful training sample selection \cite{zheng2016person,zheng2016mars}. Yet the application of the identification loss requires more training instances per ID for model convergence. For comparison, this survey presents some baseline results for both types of models. In Table \ref{table:identification_verification}, we implement the identification and verification models on the Market-1501 dataset \cite{zheng2015scalable}. All the networks use the default parameter settings, and are fine-tuned from the ImageNet \cite{deng2009imagenet} pre-trained models. Images are resized to $224\times224$ before being fed into the network. The initial learning rate is set to 0.001 and reduced by a factor of 0.1 after each epoch. Training is done after 36 epochs. We can clearly observe that the identification model outperforms the verification model, and that the residual-50 model  \cite{he2016deep} yields state-of-the-art re-ID accuracy on Market-1501 compared with recent results \cite{varior2016gated,varior2016siamese,su2016deep}.

The above-mentioned works learn deep features in an end-to-end manner, and there are alternatives that take low-level features as input. In \cite{wu2016deep}, low-level descriptors including SIFT and color histograms are aggregated into a single Fisher Vector \cite{perronnin2010improving} for each image. The hybrid network builds fully connected layers on the input Fisher vectors and enforces the linear discriminative analysis (LDA) as an objective function to produce embeddings that have low intra-class variance and high inter-class variance. Wu \emph{et al.} \cite{wu2016enhanced} propose concatenating the FC feature and a low-level feature vector, which is followed by another FC layer before the softmax loss layer. This method constrains the FC features using the hand-crafted features.
\setlength{\tabcolsep}{3.75pt}
\begin{table}[t]
\centering
\scriptsize
\caption{Statistics of some commonly used datasets \cite{gray2007evaluating,zheng2009associating,loy2009multi,Cheng2011custom,hirzer2011person,martinel2012re,li2012human,li2013locally,li2014deepreid,das2014consistent,roth14a,zheng2015scalable} for image-based re-ID.}
\begin{tabular}{l|cccccc}
\hline
Dataset& time& \#ID & \#image & \#camera & label & evaluation\\
\hline
VIPeR& 2007 & 632 & 1,264 &2 & hand&CMC\\
iLIDS& 2009  & 119 & 476 &2 & hand&CMC\\
GRID&  2009 & 250 &1,275 &8 & hand&CMC\\
CAVIAR& 2011  & 72 & 610&2 & hand&CMC\\
PRID2011&  2011 & 200 &1,134 &2 & hand&CMC\\
WARD&  2012 & 70 &4,786 &3 & hand&CMC\\
CUHK01& 2012  & 971 & 3,884&2 & hand&CMC\\
CUHK02& 2013 & 1,816 & 7,264 &10 (5 pairs) & hand&CMC\\
CUHK03& 2014 & 1,467 & 13,164 &2 & hand/DPM&CMC\\
RAiD&  2014 & 43 & 1,264 &4 & hand&CMC\\
PRID 450S &  2014 & 450 & 900 &2 & hand&CMC\\
Market-1501& 2015  & 1,501 & 32,668 &6 & hand/DPM&CMC/mAP\\
\hline
\end{tabular}
\label{table:compare_dataset}
\end{table}

\subsection{Datasets and Evaluation}
\subsubsection{Datasets}
A number of datasets for image-based re-ID have been released, and some commonly used datasets are summarized in Table \ref{table:compare_dataset}. The most tested benchmark is VIPeR. It contains 632 identities, and two images for each identity. 10 random train/test splits are used for stable performance, and each split has 316 different identities in both the training and testing sets. These datasets reflect various scenarios. For example, the GRID dataset \cite{loy2009multi} was collected in an underground station, iLIDS \cite{zheng2009associating} was captured at an airport arrival hall, and CUHK01 \cite{li2012human}, CUHK02 \cite{li2013locally}, CUHK03 \cite{li2014deepreid} and Market-1501 \cite{zheng2015scalable} were collected in a university campus. Over recent years, progress can observed in several aspects.

First, the dataset scale is increasing. Many of these datasets are relatively small in size, especially those of the early days, but recent datasets, such as CUHK03 and Market-1501, are larger. Both have over 1,000 IDs and over 10,000 bounding boxes, and both datasets provide good amount of data for training deep learning models. That said, we must admit that the current data volume is still far from satisfactory. The community is in great need of larger datasets.

Second, the bounding boxes tend to be produced by pedestrian detectors (such as DPM \cite{felzenszwalb2010object} and ACF \cite{dollar2014fast}) instead of being hand-drawn. For practical applications, it is infeasible to draw gallery bounding boxes using human labor, so detectors must be used. This may cause the bounding boxes to deviate from ideal ones.  It is shown in \cite{li2014deepreid} that using detected bounding boxes usually leads to compromised re-ID accuracy compared to hand-drawn ones due to detector errors such as misalignment. In \cite{zheng2015scalable}, a number of false detection results (on the background) are included in the gallery, which is inevitable when detectors are used. The experiments in \cite{zheng2015scalable} show that re-ID accuracy drops as more distractors are added to the gallery. As a consequence, it is beneficial for the community to study datasets with practical imperfections such as false detection and misalignment.

Third, more cameras are used during collection. For example, each identity in Market-1501 can be captured by up to 6 cameras. This design calls for metric learning methods that have good generalization ability, instead of being carefully tuned between a certain camera pair. In fact, in a city-scale camera network with $n$, the number of camera pairs is $C_n^2$, so it is prohibitive to collect annotated data from each camera and train $C_n^2$ distance metrics.

For more detailed descriptions of these datasets, we refer to survey \cite{bedagkar2014survey} and website\footnote{\url{http://robustsystems.coe.neu.edu/sites/robustsystems.coe.neu.edu/files/systems/projectpages/reiddataset.html}}.

\subsubsection{Evaluation Metrics}
When evaluating re-ID algorithms, the cumulative matching characteristics (cmc) curve is usually used. CMC represents the probability that a query identity appears in different-sized candidate lists. No matter how many ground truth matches there are in the gallery, only the first match is counted in the CMC calculation. So basically, CMC is accurate as an evaluation method only when one ground truth for each query exists. This measurement is acceptable, in practice, when people care more about returning the ground truth match in the top positions of the rank list.

For research integrity, however, when multiple ground truths exist in the gallery, Zheng \emph{et al.} \cite{zheng2015scalable} propose using the mean average precision (mAP) for evaluation. The motivation is that a perfect re-ID system should be able to return all true matches to the user. The case might be that two systems are equally competent at spotting the first ground truth, but have different retrieval recall ability. In this scenario, CMC does not have enough discriminative ability but mAP does. Therefore, mAP is used together with CMC for the Market-1501 dataset where multiple ground truths from multiple cameras exist for each query. Later, in \cite{huang2016camera,varior2016siamese,varior2016gated}, mAP results are also reported for datasets with multiple ground truths per query.
\makeatother
\begin{figure*} [t]
\centering
\subfigure[VIPeR]{\label{fig:viper}%
\includegraphics[width=2.3in]{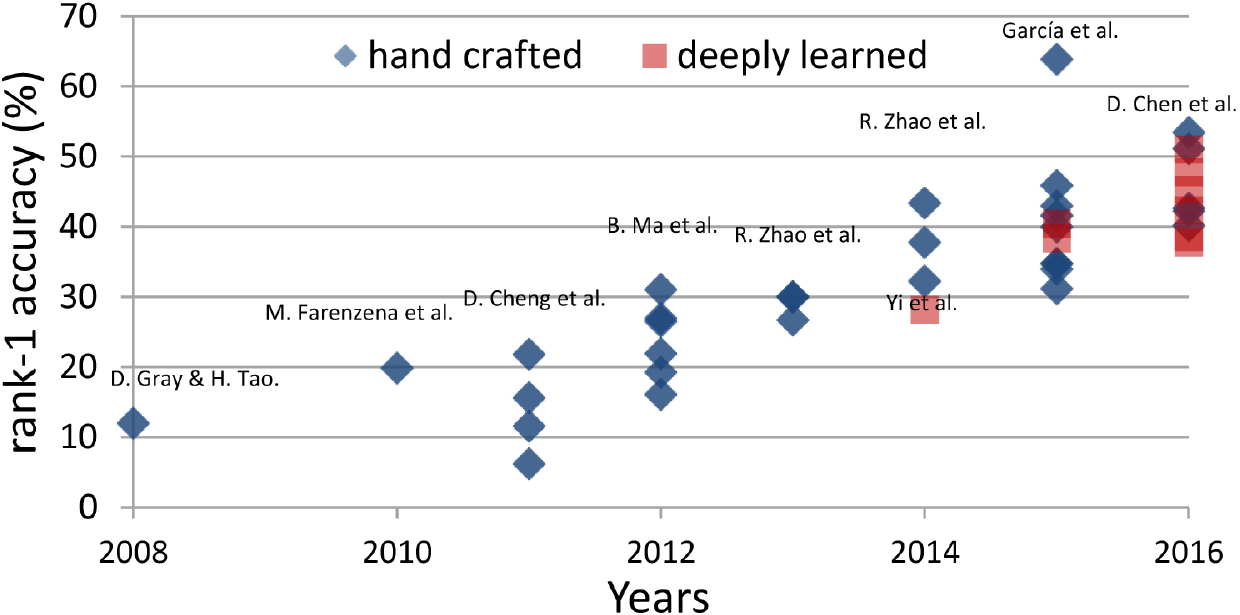}}
\hspace{0.03in}
 \subfigure[CUHK01]{\label{fig:cuhk01}%
\includegraphics[width=2.3in]{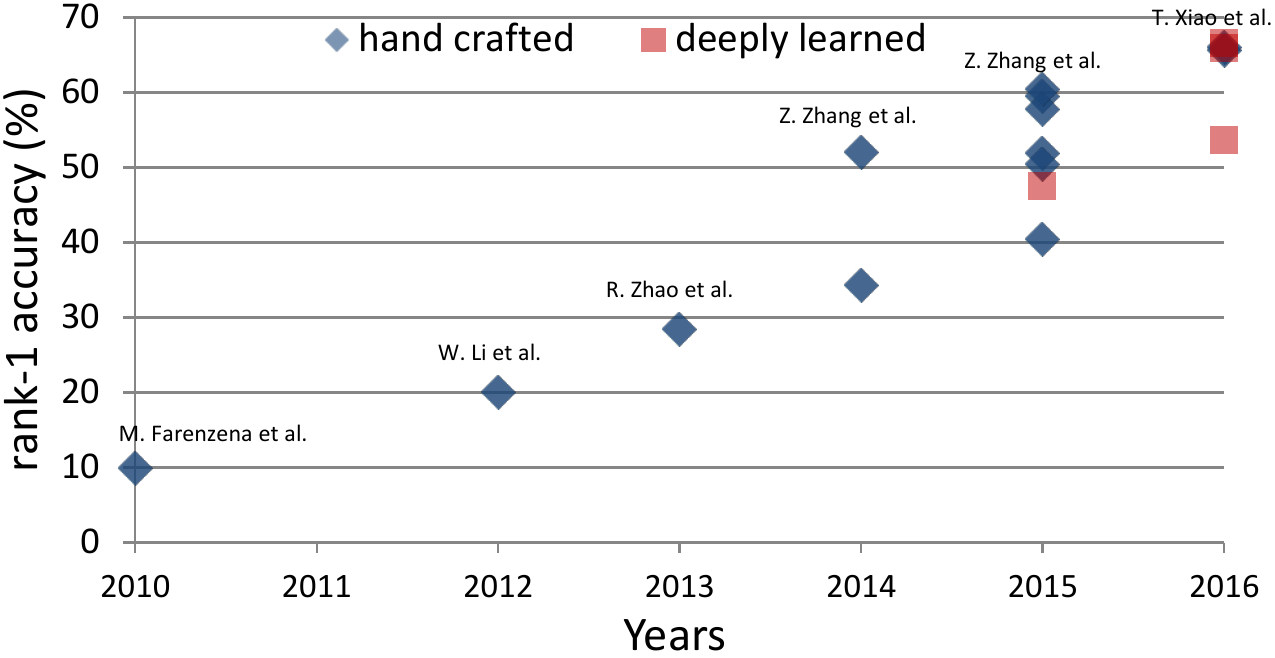}}
\hspace{0.03in}
 \subfigure[iLIDS]{\label{fig:iLIDS}%
\includegraphics[width=2.3in]{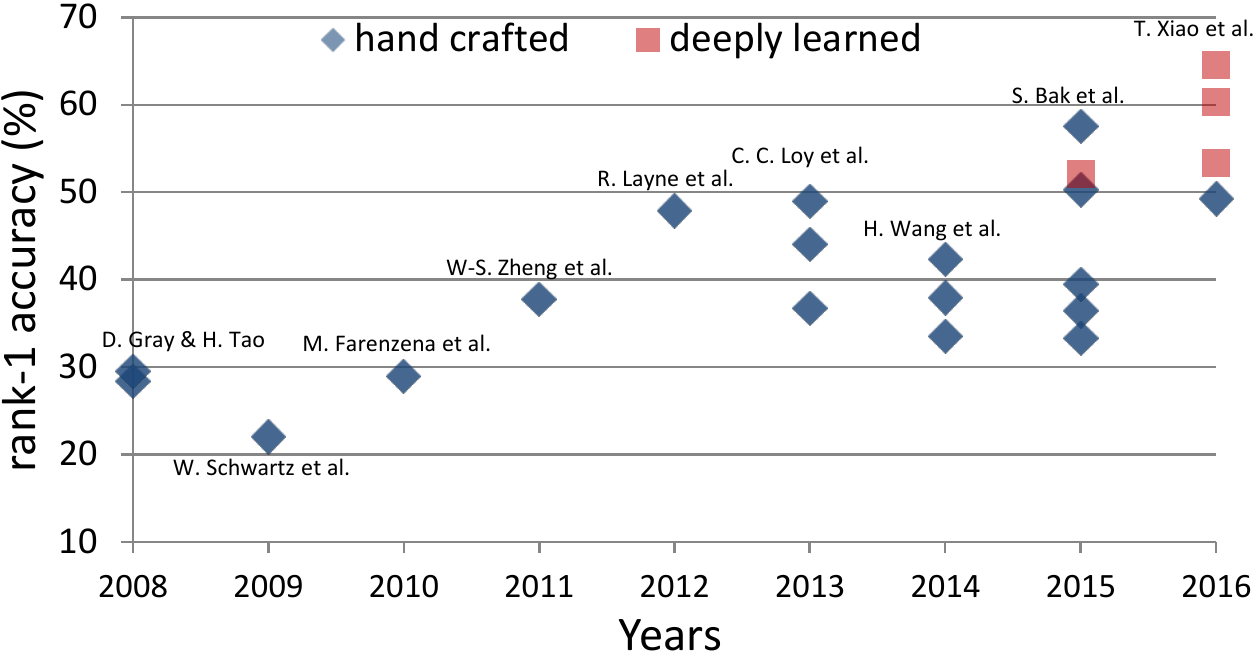}}
\subfigure[PRID 450S]{\label{fig:prid450s}%
\includegraphics[width=2.3in]{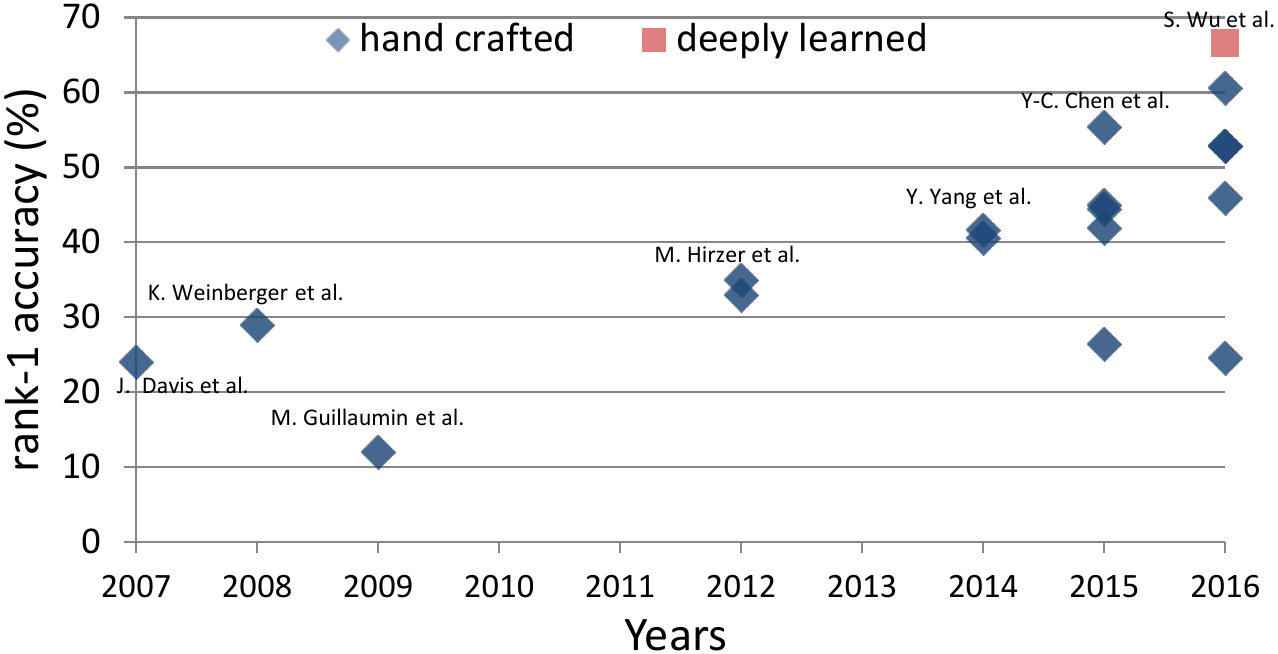}}
\hspace{0.03in}
 \subfigure[CUHK03]{\label{fig:CUHK03}%
\includegraphics[width=2.3in]{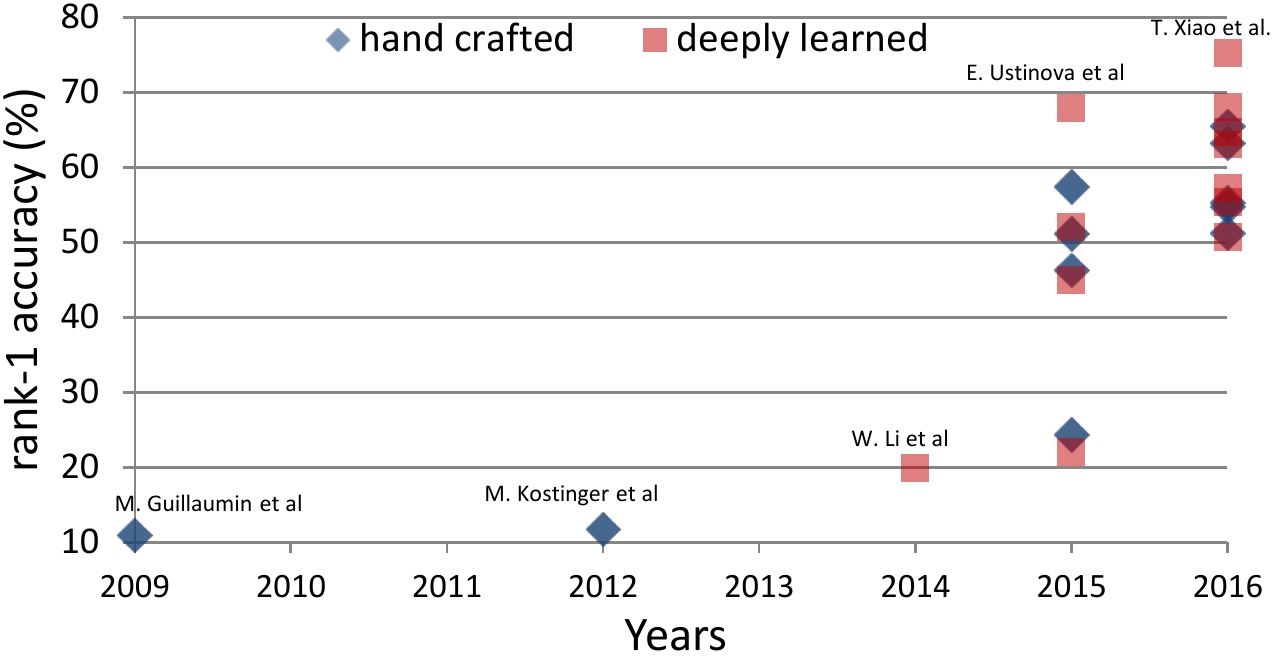}}
\hspace{0.03in}
 \subfigure[Market-1501]{\label{fig:market1501}%
\includegraphics[width=2.3in]{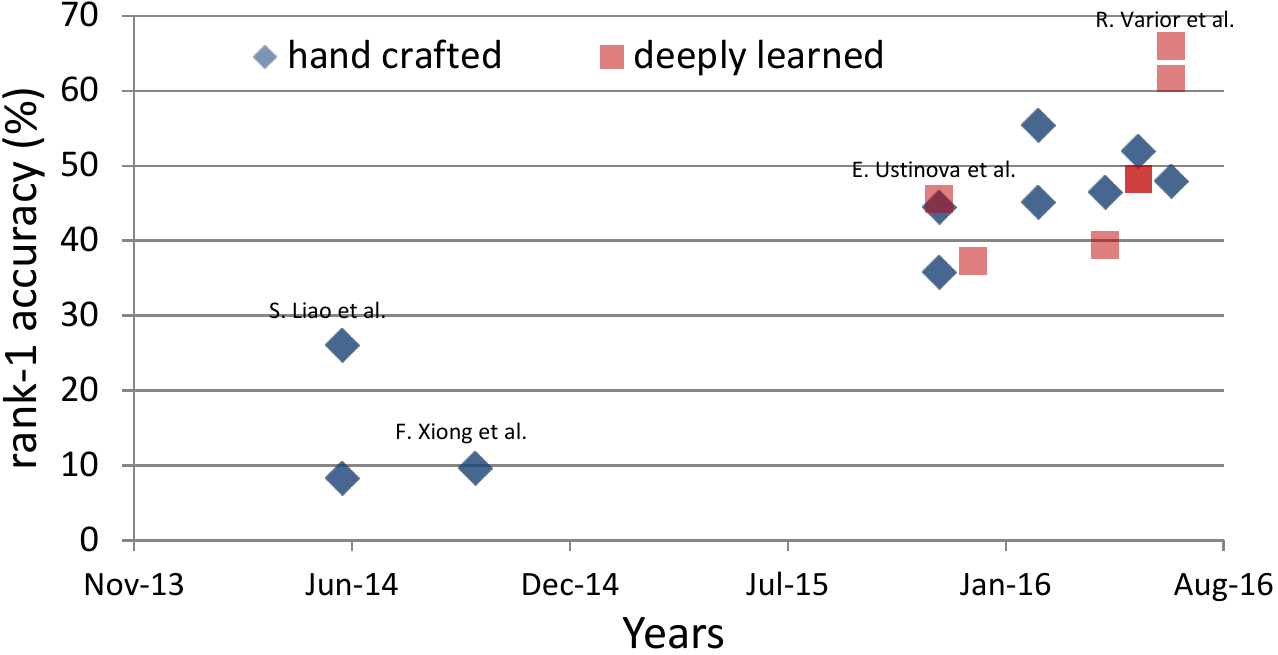}}
\caption{Person re-ID accuracy on (a) VIPeR \cite{gray2007evaluating} (b) iLIDS \cite{zheng2009associating} (c) GRID \cite{loy2009multi} (d) CUHK01 \cite{li2012human} (e) CUHK03 \cite{li2014deepreid} and (f) Market-1501 \cite{zheng2015scalable}  over the years. Results from top venues using hand-crafted or deeply learned systems are presented. For CUHK03, we record results on the detected data, and for Market-1501, results using single queries are used. Since Market-1501 was  released recently, results on this dataset are plotted according to their publication (or ArXiv) time.}
\label{fig:image-sota}
\end {figure*}
\subsubsection{Re-ID Accuracy Over the Years}
In this section, we summarize re-ID accuracy on several representative datasets over the years in Fig. \ref{fig:image-sota}. The presented datasets are VIPeR \cite{gray2007evaluating}, CUHK01 \cite{li2012human}, iLIDS \cite{zheng2009associating}, PRID 450S \cite{roth14a}, CUHK03 \cite{li2014deepreid}, and Market-1501 \cite{zheng2015scalable}. We broadly classify the current methods into two types, \emph{i.e.,} hand crafted and deeply learned. For each dataset, representative methods that report the highest re-ID accuracy in the corresponding year are shown. From these results, three major insights can be drawn.

First, a clear trend of performance improvement can be observed from the six datasets over the years. On VIPeR, CUHK01, i-LIDS, PRID 450S, CUHK03, and Market-1501, we observe a performance increase of +51.9\%, +56.7\%, +35.0\%, +42.6\%, +57.2\%, and +31.62\%, respectively.  For example, on the most studied  dataset VIPeR \cite{gray2007evaluating}, from the year 2008 to 2016, representative works \cite{gray2008viewpoint,garcia2015person,farenzena2010person,Cheng2011custom,zhao2013unsupervised,paisitkriangkrai2015learning,chen2016similarity} witness a rank-1 accuracy from 12.0\% in 2008 \cite{gray2008viewpoint} to 63.9\% in 2015 \cite{garcia2015person}, an improvement of +51.9\%. For the Market-1501 dataset, since its release in 2015, the state-of-the-art results have increased from 44.42\% \cite{zheng2015scalable} to 76.04\% \cite{varior2016gated}, an improvement of 31.62\%.

Second, with the exception of VIPeR, deep learning methods yield a new state of the art on the remaining 5 datasets. On these 5 datasets (CUHK01, i-LIDS, PRID 450S, CUHK03, and Market-1501), the performance of deep learning is superior to hand-crafted systems. On CUHK03 and Market-1501, the two largest datasets so far, we observe overwhelming advantage for deep learning \cite{xiao2016learning,varior2016gated} compared to the (also extensive) tests of hand-crafted methods. Since VIPeR is relatively small, the advantage of deep learning cannot be tested to the full; instead, a hand-crafted metric learning may be more advantageous in this setting. Considering the cases in image classification and object detection, it is highly possible that deeply learned systems will continue dominating the re-ID community over the next few years.

Third, we speculate that there is still much room for further improvement, especially when larger datasets are to be released. For example, on the Market-1501 dataset, while the best rank-1 accuracy is 65.88\% without using multiple queries \cite{varior2016gated}, mAP is quite low (39.55\%). This indicates that although it is relative easy to find the first true match (rank-1 accuracy) among a pool of 6 cameras, it is not trivial to locate the hard positives and thus achieve a high recall (mAP). On the other hand, although we seem to be able to achieve 60\% to 70\% rank-1 accuracy on these datasets, we must keep in mind that these datasets receive a very small proportion of practical usage. In fact, apart from \cite{zheng2015scalable},
 it is also reported in \cite{wang2016human} a 10-fold gallery size increase leads to a 10-fold decrease in rank-1 accuracy, resulting in  a single-digit rank-1 score even for the best-performing methods. As a consequence, considering the low mAP (re-ID recall) and the small scale of current datasets, we are more than optimistic that important breakthroughs are to be expected in image-based re-ID.

\section{Video-based Person Re-ID}\label{sec:video_reid}
In literature, person re-ID is mostly explored with single images (single shot). In recent years, video-based re-ID has become popular due to the increased data richness which induces more research possibilities. It shares a similar formulation to image-based re-ID as Eq. \ref{eq:img_reid}. Video-based re-ID replaces images $q$ and $g$ with two sets of bounding boxes $\{q_i\}_{i=1}^{n_q}$ and $\{g_j\}_{j=1}^{n_g}$, where $n_q$ and $n_g$ are the number of bounding boxes within each video sequence, respectively. As important as the bounding box features are, video-based methods pay additional attention to multi-shot matching schemes and the integration of temporal information.

\subsection{Hand-crafted Systems}
The  first two trials \cite{bazzani2010multiple,farenzena2010person} in 2010 were both hand-crafted systems. They basically use color-based descriptors and optionally employ foreground segmentation to detect the pedestrian. They use similar image features to image-based re-ID methods, where the major difference is the matching function. As mentioned in Section \ref{sec:history}, both methods commonly calculate the minimum Euclidean distance between two sets of bounding box features as the set similarity. In essence, such methods should be classified into ``multi-shot'' person re-ID, where the similarity between two sets of frames plays a critical role. This multi-shot matching strategy is adopted by later works \cite{ma2012bicov,ma2012local}. In \cite{hirzer2011person}, multiple shots are used to train a descriminative boosting model based on a set of covariance features.  In \cite{hamdoun2008person}, the SURF local feature is used to detect and describe interest points within short video sequences that are in turn indexed in the KD-tree to speed up matching. In \cite{gheissari2006person}, a spatial-temporal graph is generated to identify spatial-temporal stable regions for foreground segmentation. The the local descriptions are then calculated using a clustering method over time to improve matching performance. Cong \emph{et al.} \cite{cong2009video} employ the manifold geometric structures from video sequences to construct more compact spatial descriptors with color-based features. Karaman \emph{et al.} \cite{karaman2012identity} propose using the conditional random field (CRF) to incorporate constraints in the spatial and temporal domains. In \cite{bedagkar2012part}, colors and selected face images are used to build a model over frames that capture the characteristic appearance as well as its variations over time. Karanam \emph{et al.} \cite{karanam2015sparse} make use of multi-shots for a person and propose that the probe feature be presented as a linear combination of the same person in the gallery. Multiple shots of an identity can also be employed to enhance body part alignment. In \cite{Cheng2011custom}, in the effort to look for precise part-to-part correspondence, Cheng \emph{et al.} propose an iterative algorithm in which the fitting of the pictorial structure becomes more accurate  after each iteration  due to the improvement of part detectors. In \cite{cho2016improving}, pedestrian poses are estimated and frames with the same pose are matched with higher confidence.

The above methods typically build appearance models based on multiple shots, and a recent trend is to incorporate temporal cues in the model. Wang \emph{et al.} \cite{wang2014person} propose using spatial-temporal descriptors to re-identify pedestrians. Its features include HOG3D \cite{klaser2008spatio} and the gait energy image (GEI) \cite{man2006individual}. By designing a flow energy profile (FEP), walking cycles are detected so that frames around the local minimum/maximum are used to extract motion features. Finally, reliable spatial-temporal features are selected and matched through a discriminative video ranking model. In \cite{liu2015spatio}, Liu \emph{et al.} propose de-composing a video sequence into a series of units that represent body-actions corresponding to certain action primitives, from which Fisher vectors are extracted for the final representation of the person. Gao \emph{et al.} \cite{gao2016temporally} make use of the periodicity property of pedestrians and divide the walking cycle into several segments which are described by temporally aligned pooling. In \cite{liu2016fast}, a new spatial-temporal descriptor is proposed based on densely computed multi-directional gradients and discarding noisy motion occurring over a short period.

Distance metric learning is also important when matching videos. In \cite{zheng2012transfer}, a set verification method is proposed in which a transfer ranking is employed to tell whether the query matches one of the images belonging to the same identity. In \cite{li2013locally}, the multi-shot extension of the proposed local match model minimizes the distance of the best-matched pairs and reduces the number of cross-view transformations. In \cite{zhu2016video}, Zhu \emph{et al.} propose simultaneously learning intra- and inter-video distance metrics to make video representation more compact and to discriminate videos of different identities. You \emph{et al.} \cite{you2016top} propose the top-push distance learning method which optimizes the top-rank matching in video re-ID by selecting discriminative features.


\subsection{Deeply-learned Systems}
In video-based re-ID, the data volume is typically larger than image-based datasets, because each tracklet contains a number of frames (Table \ref{table:compare_video_dataset}).
\makeatother
\begin{figure*} [t]
\centering
\subfigure[ETHZ]{\label{fig:viper}%
\includegraphics[width=2.3in]{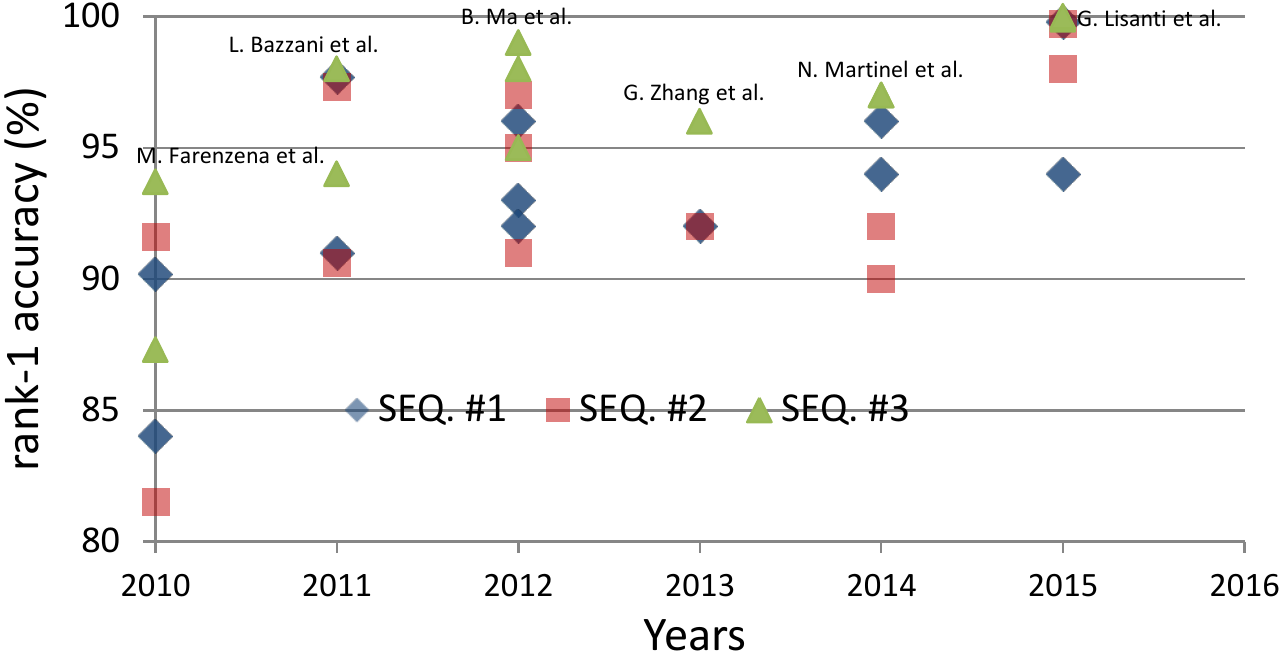}}
\hspace{0.03in}
 \subfigure[iLIDS-VID]{\label{fig:cuhk01}%
\includegraphics[width=2.3in]{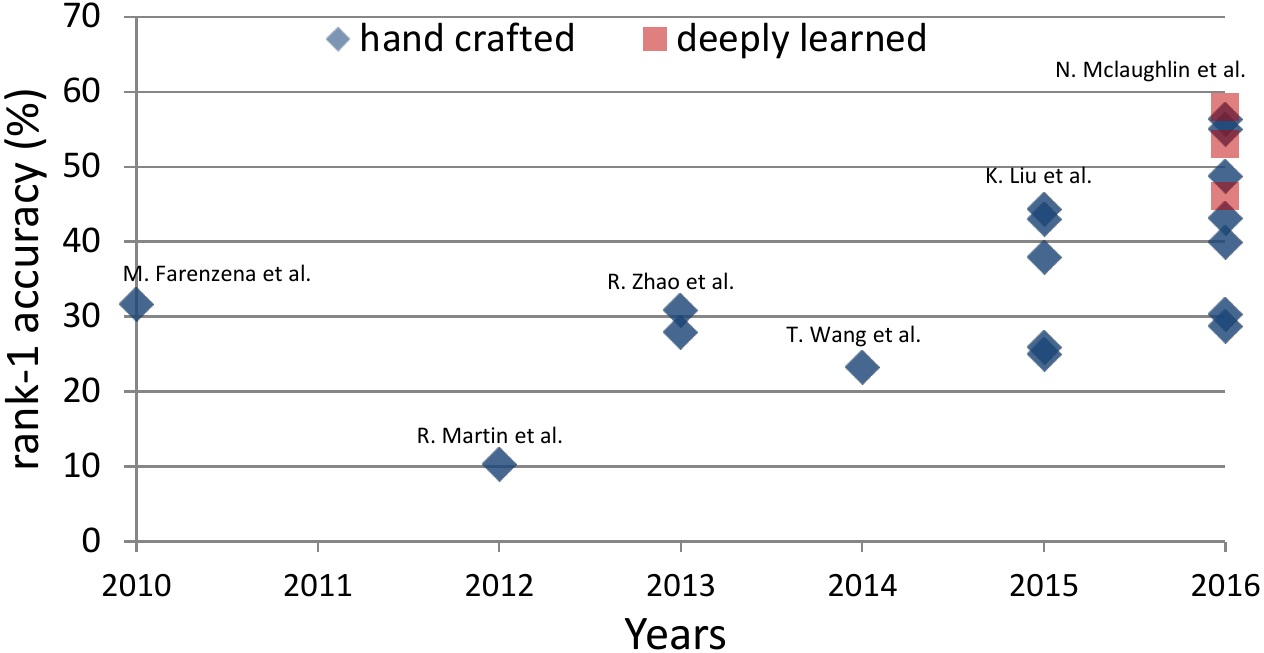}}
\hspace{0.03in}
 \subfigure[PRID-2011]{\label{fig:iLIDS}%
\includegraphics[width=2.3in]{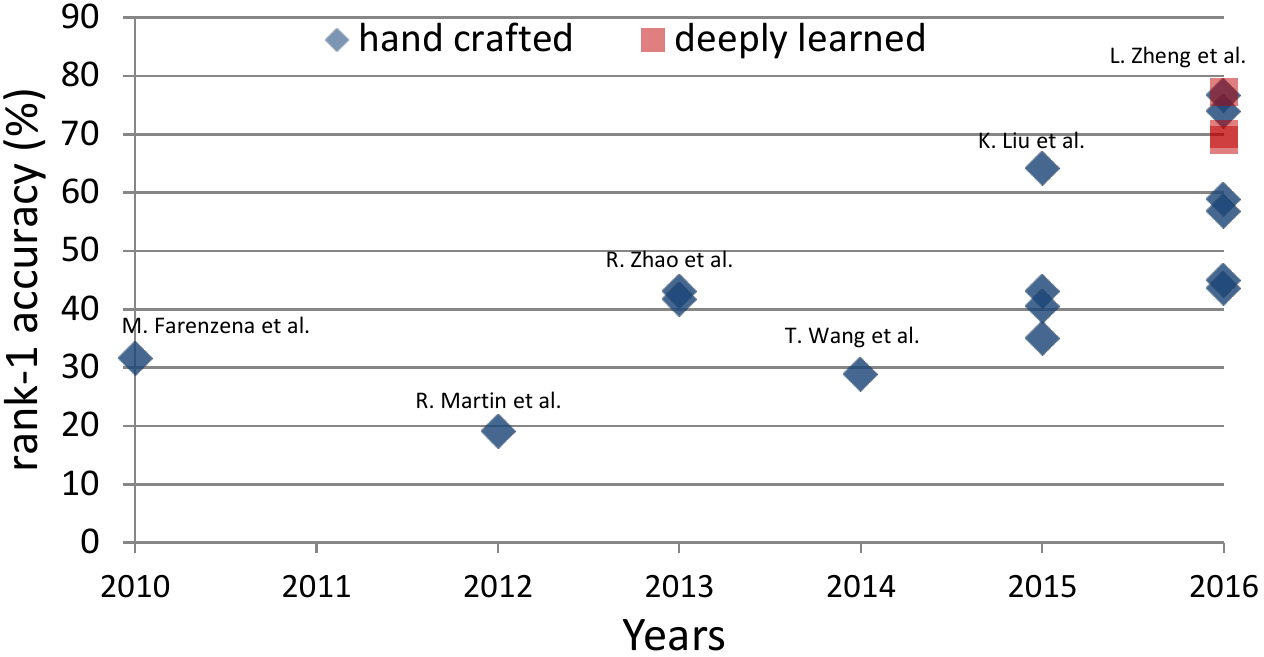}}
\caption{Video-based person re-ID accuracy on (a) ETH sequence 1 \cite{gray2007evaluating} (b) PRID-2011 \cite{hirzer2011person} and (c) iLIDS-VID \cite{wang2014person} over the years. Results from top venues using hand-crafted or deeply learned systems are presented. For ETHZ, we report results obtained by 5 images per video sequence, and state-of-the-art results on SEQ. \#1, SEQ. \#2, and SEQ. \#3 are drawn. }
\label{fig:video-sota}
\end {figure*}

A basic difference between video-based and image-based re-ID is that with multiple images for each matching unit (video sequence), either a multi-match strategy or a single-match strategy after video pooling should be employed. The multi-match strategy is used in older works \cite{bazzani2010multiple,farenzena2010person}, which induces higher computational cost and may be problematic on large datasets. On the other hand, pooling-based methods aggregates frame-level features into a global vector, which has better scalability. As a consequence, current video-based re-ID methods typically involve the pooling step. This step can be max/average pooling as \cite{zheng2016mars,mclaughlin2016recurrent}, or learned by a fully connected layer \cite{Yan2016person}. In Zheng \emph{et al.}'s system \cite{zheng2016mars}, temporal information is not explicitly captured; instead, frames of an identity are viewed as its training samples to train a classification CNN model with softmax loss. Frame features are aggregated by max pooling which yield competitive accuracy on three datasets.  These methods are proven to be effective, and yet there is plenty of space for improvement. With respect to this point, the re-ID community can borrow ideas from the community of action/event recognition. For example, Xu \emph{et al.} \cite{xu2015discriminative} propose aggregating the column features in the 5th convolutional layer of CaffeNet into Fisher vectors \cite{perronnin2010improving} or VLAD \cite{jegou2010aggregating}, in direct CNN feature transfer. Fernando \emph{et al.} \cite{fernando2016rank} propose a learning-to-rank model to capture how frame features evolve over time in a video, which yields video descriptors of video-wide temporal dynamics. Wang \emph{et al.} \cite{wang2015temporal} embed a multi-level encoding layer into the CNN model and produce video descriptors of varying sequence lengths.

Another good practice consists of injecting temporal information in the final representation. In hand-crafted systems, Wang \emph{et al.} \cite{wang2014person} and Liu \emph{et al.} \cite{liu2015spatio} use pure spatial-temporal features on the iLIDS-VID and PRID-2011 datasets and report competitive accuracy. In \cite{zheng2016mars}, however, it is shown that the spatial-temporal features are not sufficiently discriminative on the MARS dataset, because many pedestrians share similar waling motion under the same camera, and because motion feature of the same person can be distinct in different cameras. The point made in \cite{zheng2016mars} is that appearance features are critical in a large-scale video re-ID system. That said, this survey calls for attention to the recent works of \cite{Yan2016person,Wu2016deeprecurrent,mclaughlin2016recurrent}, in which appearance features (\emph{e.g.,} CNN, color and LBP) are used as the starting point to be fed into RNN networks to capture the time flow between frames. In \cite{mclaughlin2016recurrent}, features are extracted from consecutive video frames through a CNN model, and then fed through a recurrent final layer, so that information flow between time-steps is allowed. The features are then combined using max or average pooling to yield an appearance feature for the video. All these structures are incorporated into a siamese network. A similar architecture is used in \cite{Wu2016deeprecurrent}. Their difference is two-fold. First, a particular type of RNN, the Gated Recurrent Unit (GRU) is used in \cite{Wu2016deeprecurrent}. Second, an identification loss is adopted in \cite{mclaughlin2016recurrent}, which is beneficial for loss convergence and performance improvement. While the two works \cite{mclaughlin2016recurrent,Wu2016deeprecurrent} employ the siamese network for loss computation, Yan \emph{et al.} \cite{Yan2016person} and Zheng \emph{et al.} \cite{zheng2016mars} use the identification model which classifies each input video into their respective identities. In \cite{Yan2016person}, hand-crafted low-level features such as color and LBP are fed into several LSTMs and the LSTM outputs are connected to a softmax layer. In action recognition, Wu \emph{et al.} \cite{wu2015modeling} propose extracting both appearance and spatial-temporal features from a video and build a hybrid network to fuse the two types of features. In this survey, we note that perhaps the discriminative combination of appearance and spatial-temporal models is an effective solution in future video re-ID research.

\setlength{\tabcolsep}{2.75pt}
\begin{table}[t]
\centering
\scriptsize
\caption{Statistics of some currently available datasets \cite{ess2007depth,baltieri20113dpes,hirzer2011person,wang2014person,zheng2016mars} for video-based re-ID.}
\begin{tabular}{l|ccccccc}
\hline
Dataset& time& \#ID &\#track & \#bbox & \#cam. & label & evaluation\\
\hline
ETHZ& 2007 & 148 &148 & 8,580 & 1 & hand&CMC\\
3DPES& 2011  & 200& 1,000 & 200k&8& hand &CMC\\
PRID-2011&  2011 & 200 &400& 40k&2 & hand&CMC\\
iLIDS-VID& 2014  & 300&600 & 44k &2 & hand&CMC\\
MARS& 2016  & 1261&20,715 & 1M &6 & DPM\&GMMCP&mAP\&CMC\\
\hline
\end{tabular}
\label{table:compare_video_dataset}
\end{table}
\subsection{Datasets and Evaluation}
Several multi-shot re-ID datasets exist, \emph{e.g.,} ETH \cite{ess2007depth}, 3DPES \cite{baltieri20113dpes}, PRID-2011 \cite{hirzer2011person}, iLIDS-VID \cite{wang2014person}, and MARS \cite{zheng2016mars}. Some statistics of these datasets are summarized in Table \ref{table:compare_video_dataset}. The ETH dataset uses a single moving camera. It contains three sequences and multiple images from each sequence are provided. This dataset is relatively easy and the re-ID accuracy of the multi-shot scenario is nearly 100\% \cite{lisanti2015person}. The 3DPeS dataset is collected with 8 non-overlapping outdoor cameras. Although the videos are released, this dataset is typically used for single-shot re-ID. PRID-2011 and iLIDS-VID are similar in that both datasets were captured by 2 cameras and each identity has 2 video sequences. iLIDS-VID has 300 identities captured under indoor scenes. PRID-2011 has 385 and 749 identities for each outdoor camera, respectively, and in this dataset 200 identities are observed in both cameras. During testing, 178 identities are used for PRID-2011 following the proposal by \cite{wang2014person}. It is generally believed that iLIDS-VID is more challenging than PRID-2011 due to extremely heavy occlusion. The MARS dataset \cite{zheng2016mars} was recently released which is a large-scale video re-ID dataset containing 1,261 identities in over 20,000 video sequences. It is produced using the DPM detector \cite{felzenszwalb2010object} and the GMMCP tracker \cite{dehghan2015gmmcp}. Due to its recent release, we have not provided an extensive summary of results for the MARS dataset. Figure \ref{fig:video-sota} presents the evaluation of the state-of-the-art results on three representative video (multi-shot) re-ID datasets, \emph{i.e.,} ETHZ, iLIDS-VID, PRID-2011. Two major conclusions are drawn:

First, the ETHZ dataset has reached its performance saturation. In 2015, Lisanti \emph{et al.} \cite{lisanti2015person} and Martinel \emph{et al.} \cite{martinel2015re} report rank-1 accuracies approximating 100\%. In \cite{lisanti2015person}, using 5 images per sequence, the rank-1 accuracy of ETHZ sequence 1, 2, and 3 is 99.8\%, 99.7\%, and 99.9\%, respectively. Results with 10 frames per sequence is higher, achieving 100\% \cite{lisanti2015person,martinel2015re}. The primary reason is that the ETHZ dataset has relatively fewer identities, and the image variance is low due to the use of only one moving camera. This may be the first re-ID dataset to almost accomplish its initial objectives.

Second, active video re-ID research is still being conducted on the iLIDS-VID and PRID-2011 datasets. Since their introduction, we observe clear improvement of their rank-1 accuracy (including the ETHZ dataset). For iLIDS-VID, Wang \emph{et al.} \cite{wang2014person} report a rank-1 accuracy of 23.3\%, and an absolute improvement of 34.7\% can be seen when compared to McLaughlin \emph{et al.} \cite{mclaughlin2016recurrent}. On PRID-2011, Wang \emph{et al.} \cite{wang2014person} report a rank-1 accuracy = 19.0\%, and two years later, Zheng \emph{et al.} \cite{zheng2016mars} improve this score by 58.3\% using the max pooling of CNN features fine-tuned on the MARS dataset.

Third, deep learning methods are producing overwhelmingly superior accuracy in video-based re-ID. On both the iLIDS-VID and PRID-2011 datasets, the best performing methods are based on the convolutional neural network with optional insertion of a recurrent neural network \cite{mclaughlin2016recurrent,zheng2016mars}. Compared to image-based re-ID, the amount of training data is clearly larger in video re-ID. MARS provides over 500k training frames, compared to 13k in the Market-1501 dataset \cite{zheng2015scalable}, from which MARS was extended. With these training data, it is feasible to train discriminative networks not only for video-based re-ID, but also for image-based datasets. We also note that, while the rank-1 accuracy on the MARS dataset reaches 68.3\%, its mAP is still relatively low (49.3\%), and when evaluating the performance of each camera pair, performance is further lowered. As a consequence, we believe that the research of video re-ID still has good potential for improvement.

\section{Future: Detection, Tracking and Person Re-ID}\label{sec:det_track_reid}
\subsection{Previous Works}
Although person re-ID originates from multi-camera tracking, it is now studied as an independent research topic. In this survey, we view re-ID as an important future direction that will join pedestrian detection and tracking as a scenario, but in a more independent role. Specifically, we consider an end-to-end re-ID system\footnote{Here, ``end-to-end'' means spotting a query person from raw videos.} that takes raw videos as input and integrates pedestrian detection and tracking, along with re-identification.

Until recently, most re-ID works are based on two assumptions: first, that the gallery of pedestrian bounding boxes is given; second, that the bounding boxes are hand-drawn, \emph{i.e.,} with perfect detection quality. However, in practice, these two assumptions do not hold. On the one hand, the gallery size varies with the detector threshold. A lower threshold produces more bounding boxes (a larger gallery, higher recall, and lower precision), and vice versa. When the detection recall/precision undergoes changes due to different thresholds, re-ID accuracy does not remain stable. On the other hand, when pedestrian detectors are used, detection errors typically exist with the bounding boxes, such as misalignment, miss-detection, and false alarms. Moreover, when pedestrian trackers are used, tracking errors may lead to outlier frames within a tracklet, \emph{i.e.,} background or pedestrians with different identities. So the quality of pedestrian detection and tracking may have direct influence on re-ID accuracy, which has been rarely discussed in the re-ID community. In the following, we will review the several works devoted to this direction.

\begin{figure*}[t]
  \centering
  \includegraphics[width=7.2in]{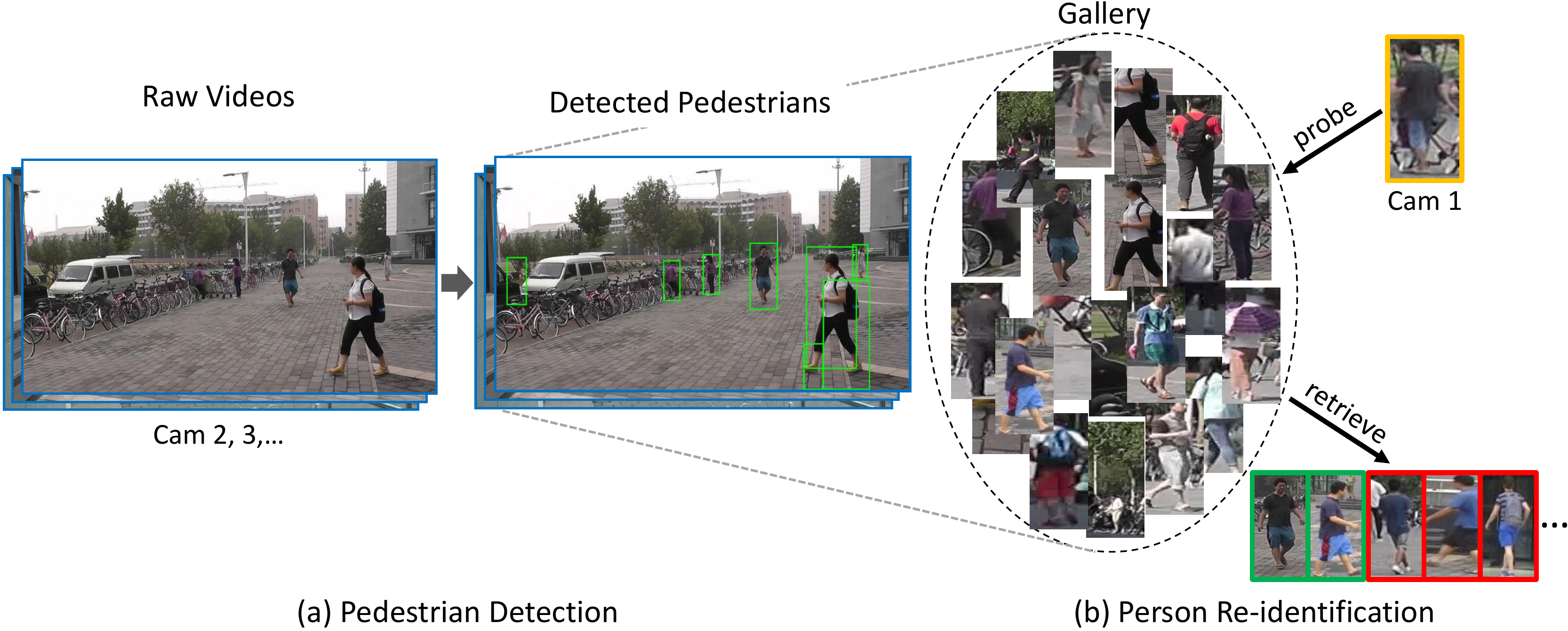}\\
  \caption{An end-to-end person re-ID system that includes pedestrian detection and re-identification. }\label{fig:pipeline}
\end{figure*}
\begin{figure}[t]
  \centering
  \includegraphics[width=3.5in]{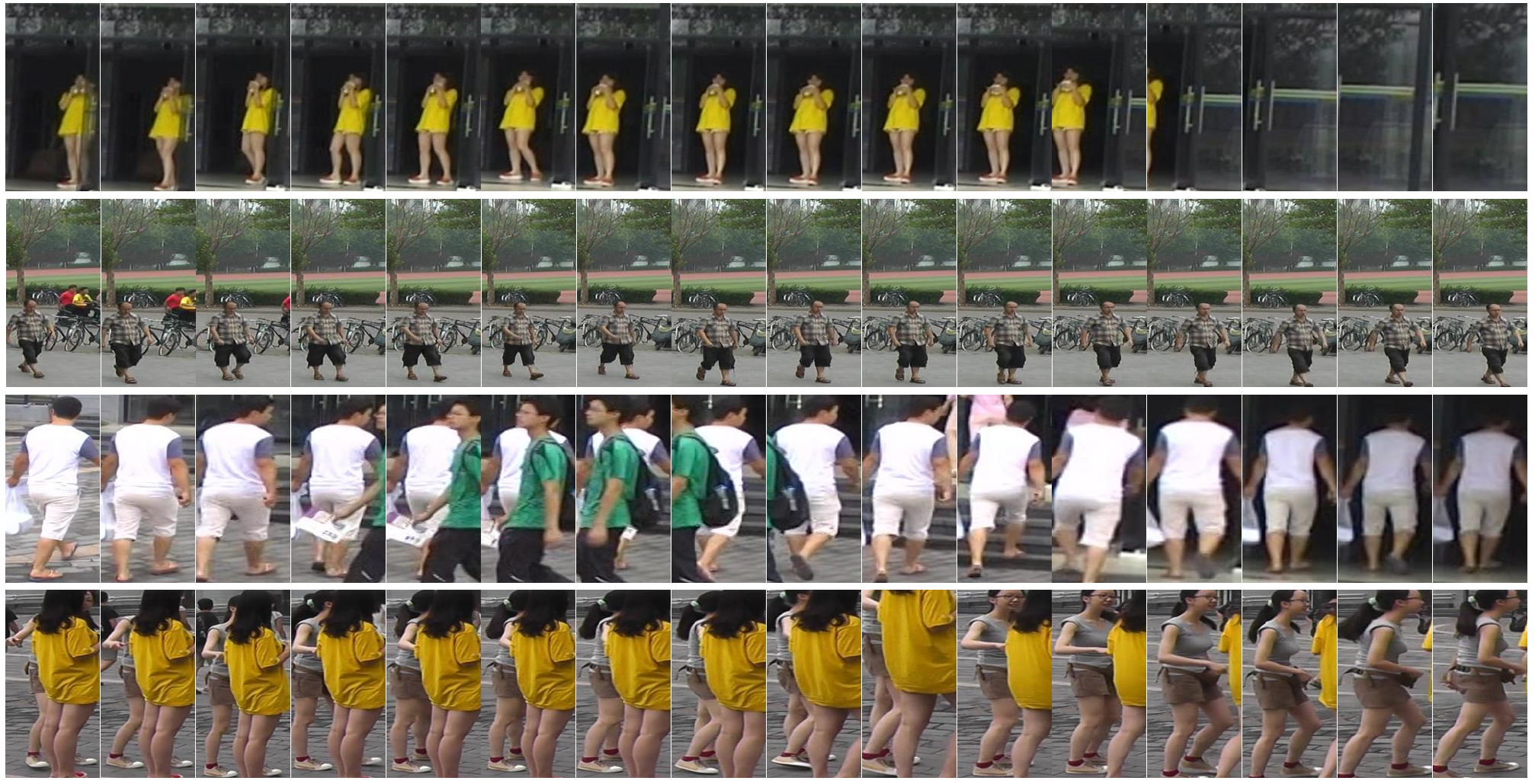}\\
  \caption{Tracking errors/artifacts in the MARS dataset \cite{zheng2016mars}. Each row represents a tracklet generated by the DPM detection + GMMCP tracker \cite{dehghan2015gmmcp}. First row: detection error and tracking error; second row: detection error; third row: occlusions in tracking; last row: tracking error. }\label{fig:track_error}
\end{figure}

In  initial attempts to address the second problem, several datasets, \emph{i.e.,} CUHK03 \cite{li2014deepreid}, Market-1501 \cite{zheng2015scalable}, and MARS \cite{zheng2016mars}, were introduced. These datasets do not assume perfect detection/tracking outputs and are a step closer to practical applications. For example, Li \emph{et al.} \cite{li2014deepreid} show that on CUHK03, re-ID accuracy using the detected bounding boxes is lower than that obtained with hand-drawn bounding boxes.  Later works also report this observation \cite{zhang2016sample,zhang2016learning}. 
These findings are closely related to practical applications. On MARS, tracking errors (Fig. \ref{fig:track_error}) as well as detection errors are presented, but it remains unknown how tracking errors will affect re-ID accuracy.

Despite the fact that the datasets make progress by introducing detection/tracking errors, they do not evaluate explicitly how detection/tracking affects re-ID, which provides critical insights into how to select detectors/trackers among the vast number of existing works in an end-to-end re-ID system. To our knowledge, the first work on end-to-end person re-ID was proposed by Xu \emph{et al.} \cite{xu2014person} in 2014. They use the term ``commonness'' to describe how an image bounding box resembles a pedestrian, and the term ``uniqueness'' to indicate the similarity between the gallery bounding box and the query. Commonness and uniqueness are fused by their product in an exponential function. This method works by eliminating the impact of false background detections. Although Xu \emph{et al.} \cite{xu2014person} considers the impact of detection on re-ID, its limitation is a lack of comprehensive benchmarking and consideration of the dynamic issue of the gallery.

In 2016, Xiao \emph{et al.} \cite{xiao2016end} and Zheng \emph{et al.} \cite{zheng2016person} simultaneously introduce an end-to-end re-ID system based on large-scale datasets. The two works take raw video frames and a query bounding box as input (Fig. \ref{fig:pipeline}). One is required to first perform pedestrian detection on the raw frames, and the resulting bounding boxes form the re-ID gallery. Then, classic re-ID is leveraged. This process, called ``person search'' in \cite{xu2014person,xiao2016end}, is no longer restricted to re-ID (Fig. \ref{fig:pipeline}(b)): it pays equal attention to the detection module (Fig. \ref{fig:pipeline}(a)). A very important aspect of this pipeline is that a better pedestrian detector tends to produce higher re-ID accuracy, given the same set of re-ID feature. In \cite{zheng2016person,xiao2016end}, extensive baselines are implemented on the person re-identification in the wild (PRW), and the large-scale person search (LSPS) datasets, respectively and this argument usually holds. Another interesting topic is whether pedestrian detection helps person re-ID. In \cite{xu2014person,zheng2016person}, detection confidence is integrated in the final re-ID scores. In \cite{xiao2016end}, pedestrian detection and re-ID are jointly considered in a CNN model which resembles faster R-CNN \cite{ren2015faster}, while in \cite{zheng2016person}, the ID-discriminative embedding (IDE) is shown to be superior when fined-tuned on a CNN model pre-trained on the R-CNN model \cite{girshick2014rich} for pedestrian detection. These methods provide initial insights on how weakly labeled detection data helps improve re-ID accuracy.

Nevertheless, in the so-called ``end-to-end'' systems \cite{xu2014person,xiao2016end,zheng2016person}, pedestrian tracking is not mentioned nor have we known any existing works/datasets addressing the influence of tracking on re-ID. This work views it as an ``ultimate'' goal to integrate detection, tracking, and retrieval into one framework, and evaluate the impact of each module on the overall re-ID performance. This survey therefore calls for large-scale datasets that provide bounding box annotations to be used for the three tasks.

\makeatother
\begin{figure*} [t]
\centering
\subfigure[BoW, r1, IoU = 0.5]{\label{fig:viper}%
\includegraphics[width=1.7in]{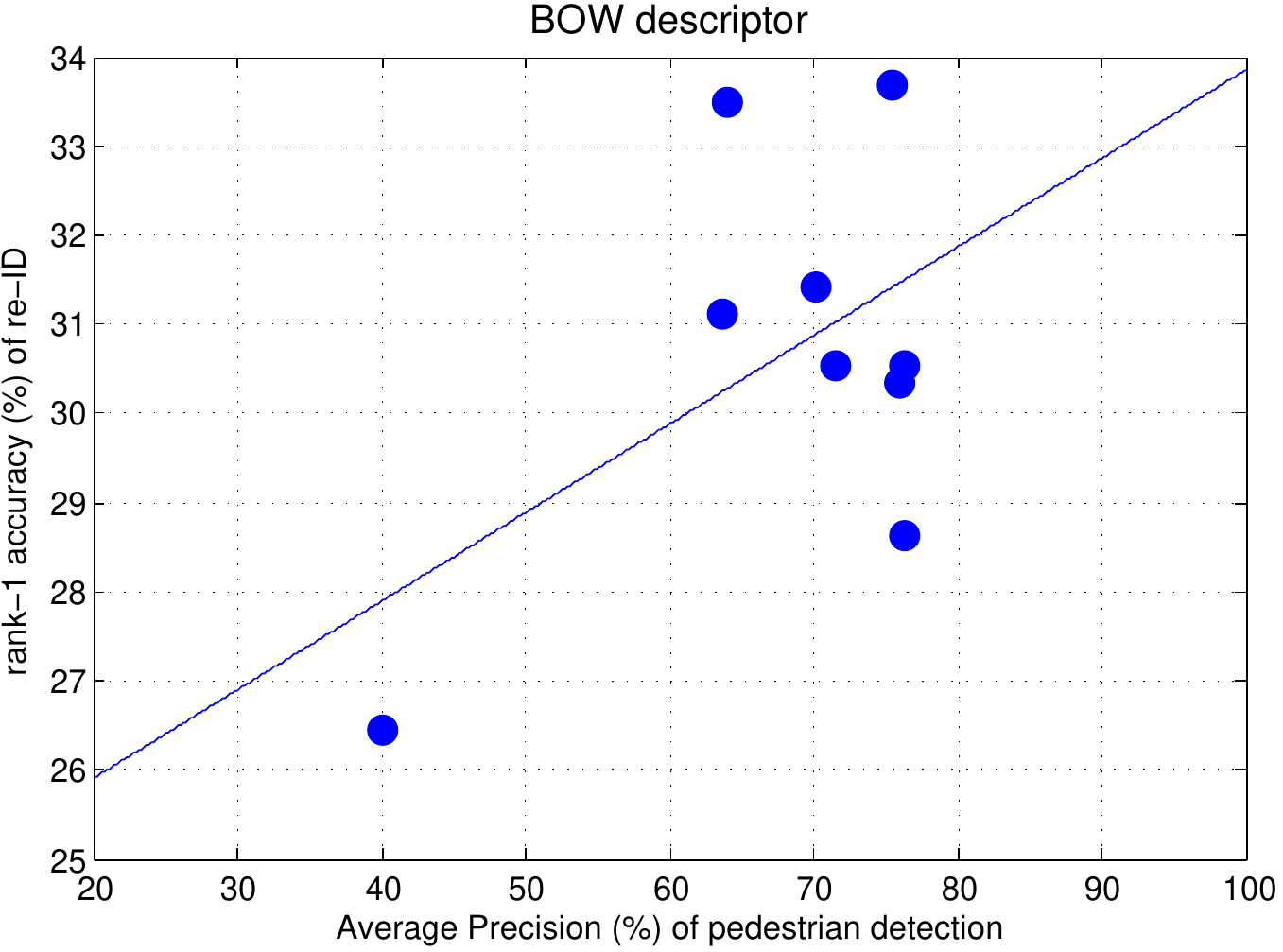}}
 \subfigure[BoW, r1, IoU = 0.7]{\label{fig:cuhk01}%
\includegraphics[width=1.7in]{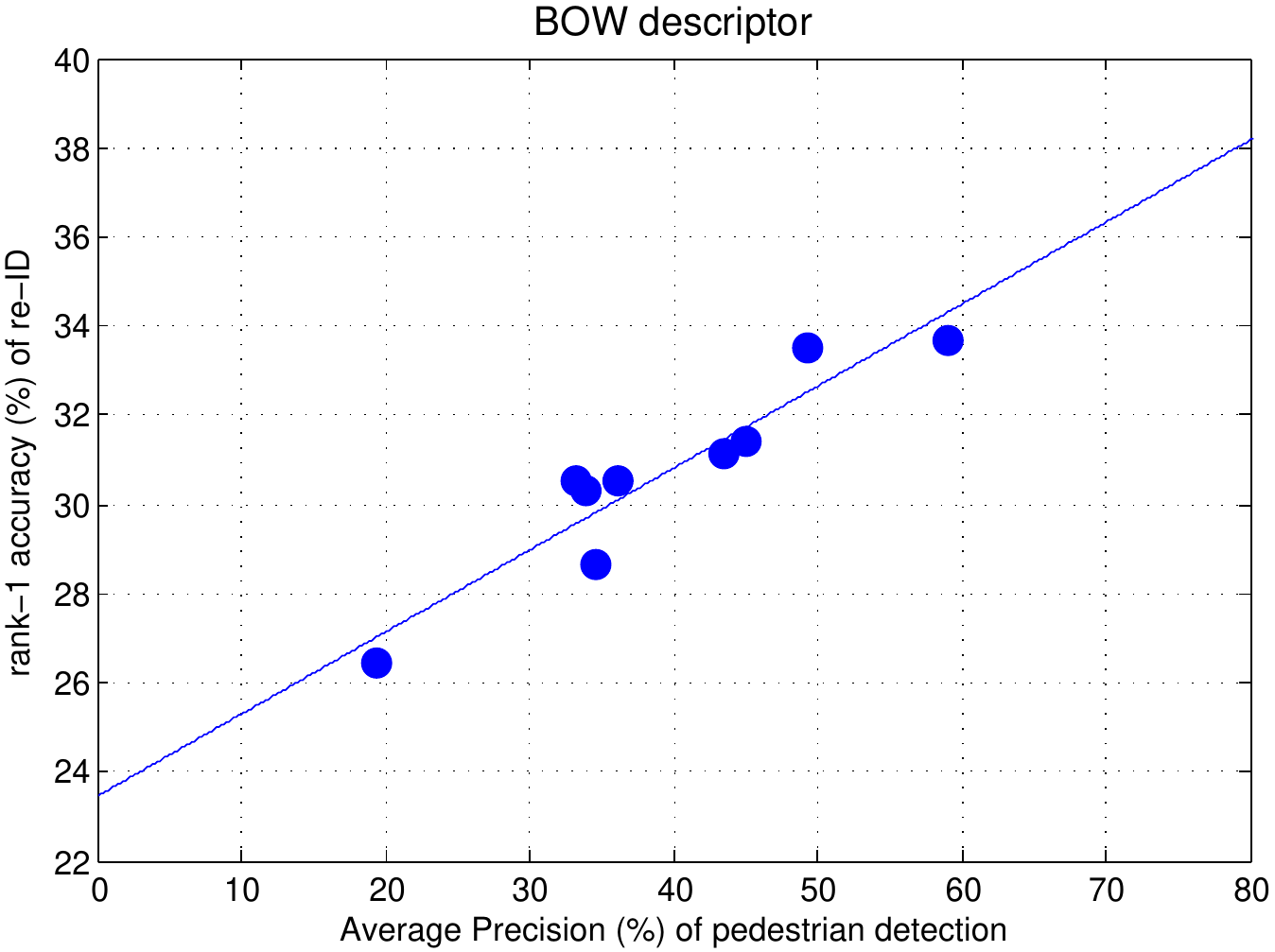}}
 \subfigure[BoW, mAP, IoU = 0.5]{\label{fig:iLIDS}%
\includegraphics[width=1.7in]{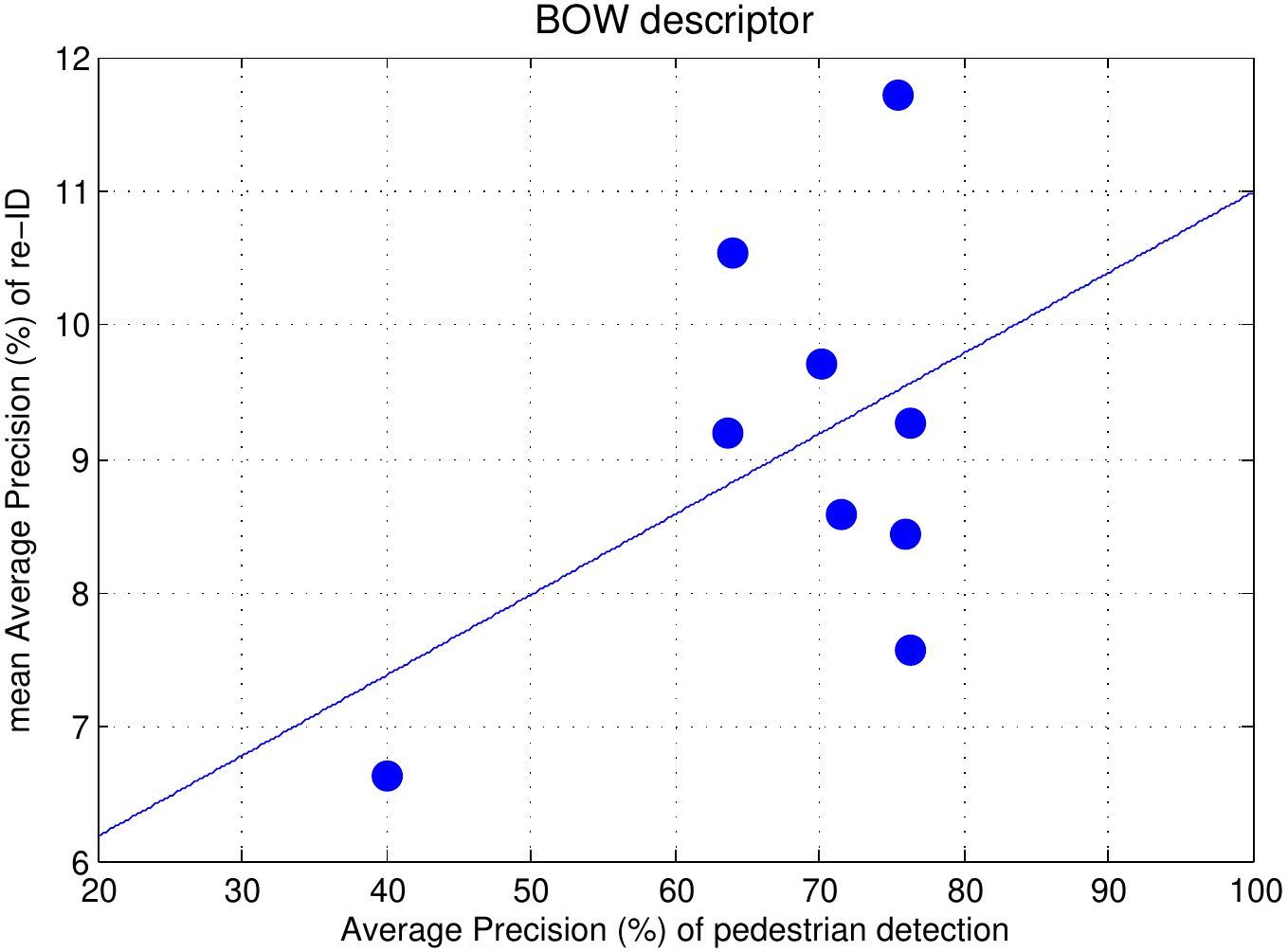}}
\subfigure[BoW, mAP, IoU = 0.7]{\label{fig:prid450s}%
\includegraphics[width=1.7in]{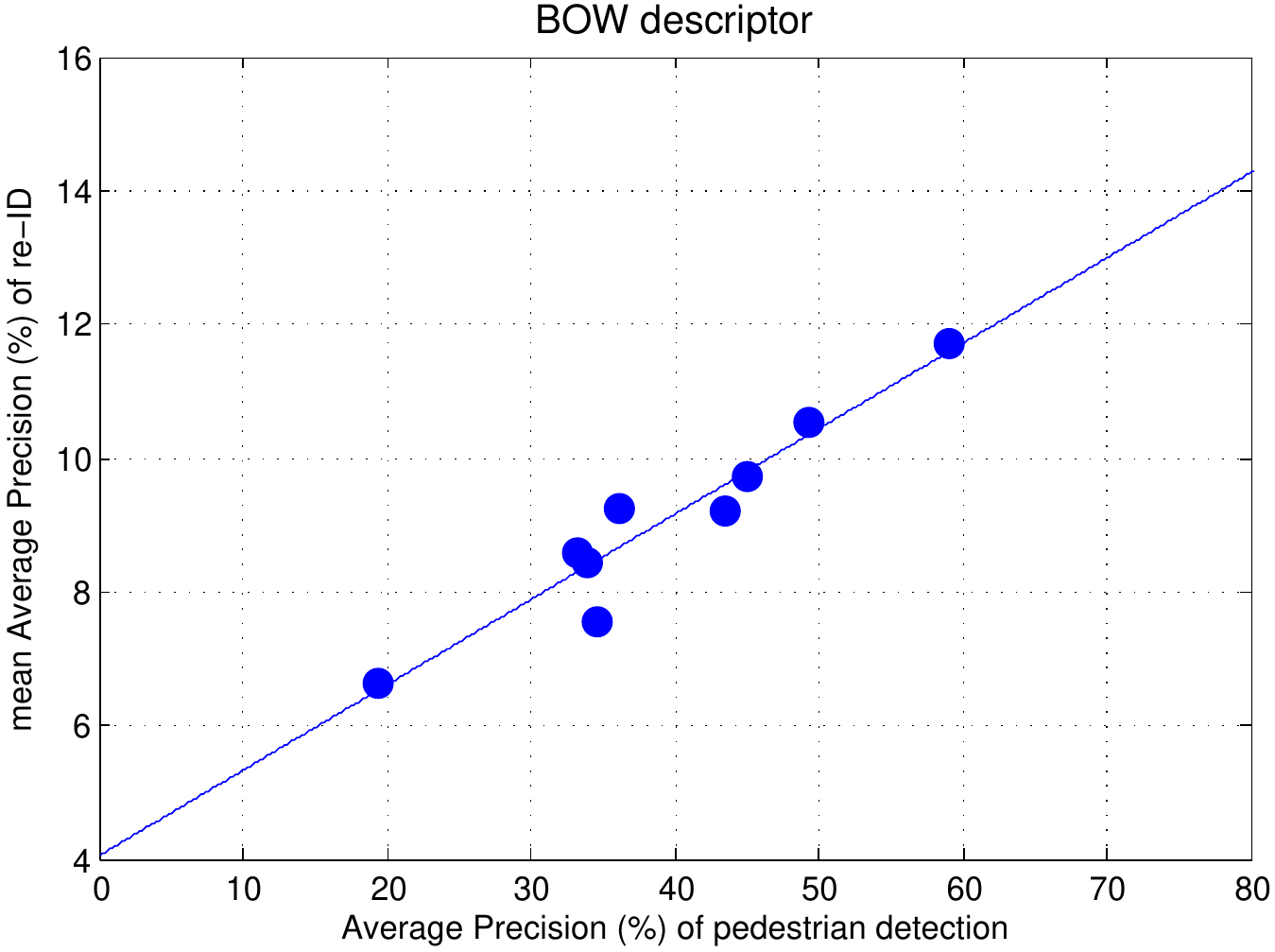}}

\subfigure[LOMO, r1, IoU = 0.5]{\label{fig:viper}%
\includegraphics[width=1.7in]{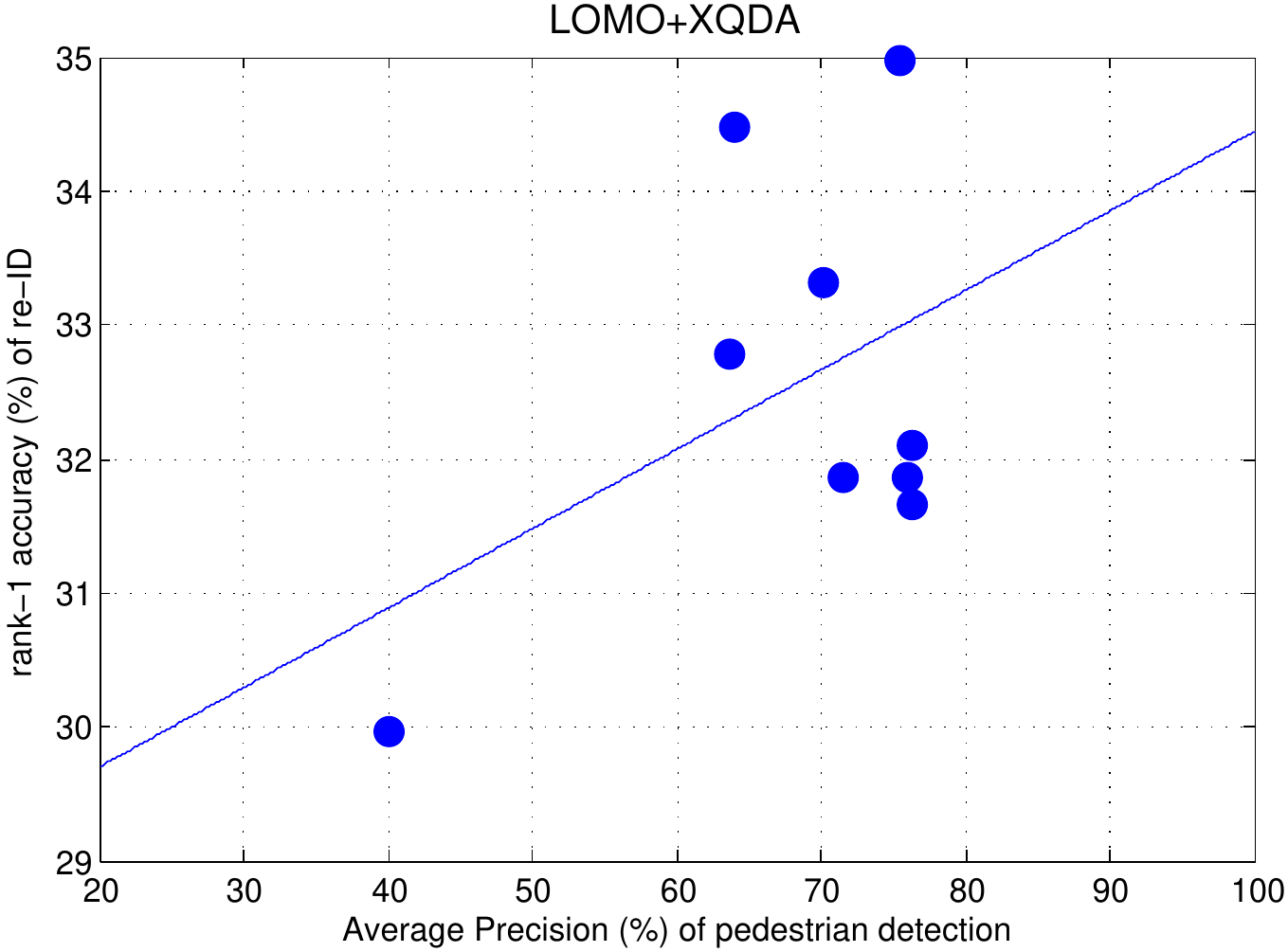}}
 \subfigure[LOMO, r1, IoU = 0.7]{\label{fig:cuhk01}%
\includegraphics[width=1.7in]{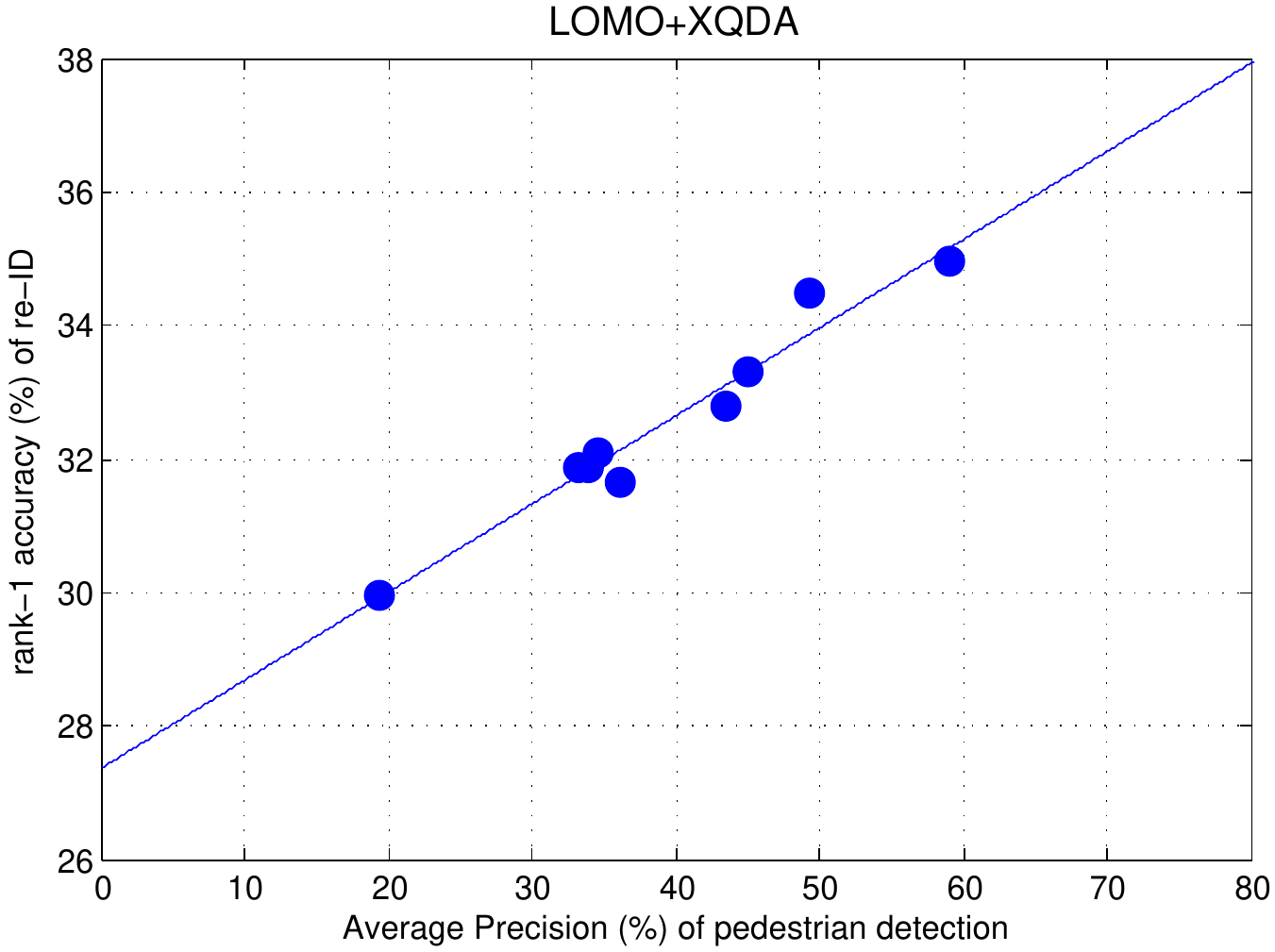}}
 \subfigure[LOMO, mAP, IoU = 0.5]{\label{fig:iLIDS}%
\includegraphics[width=1.7in]{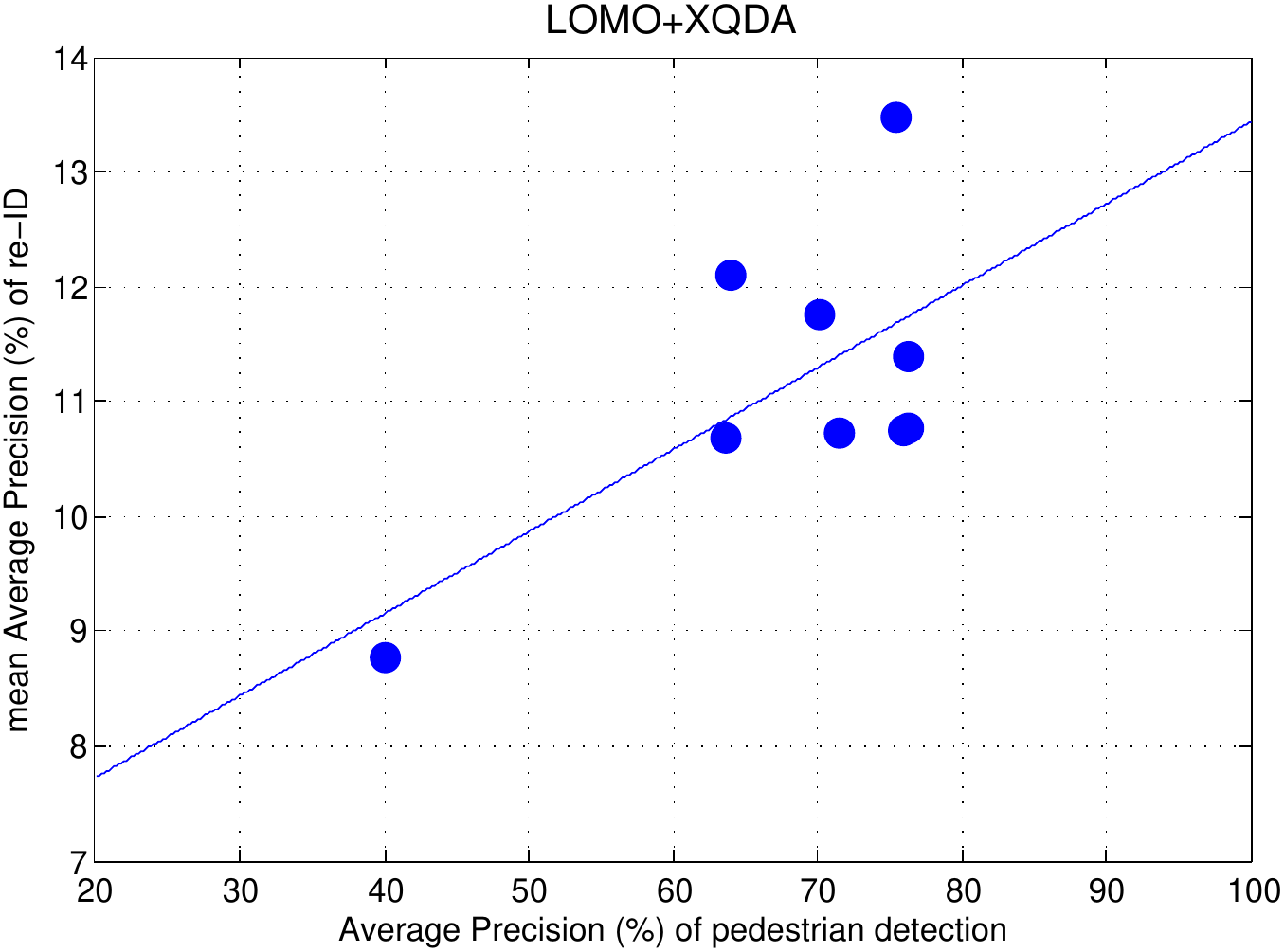}}
\subfigure[LOMO, mAP, IoU = 0.7]{\label{fig:prid450s}%
\includegraphics[width=1.7in]{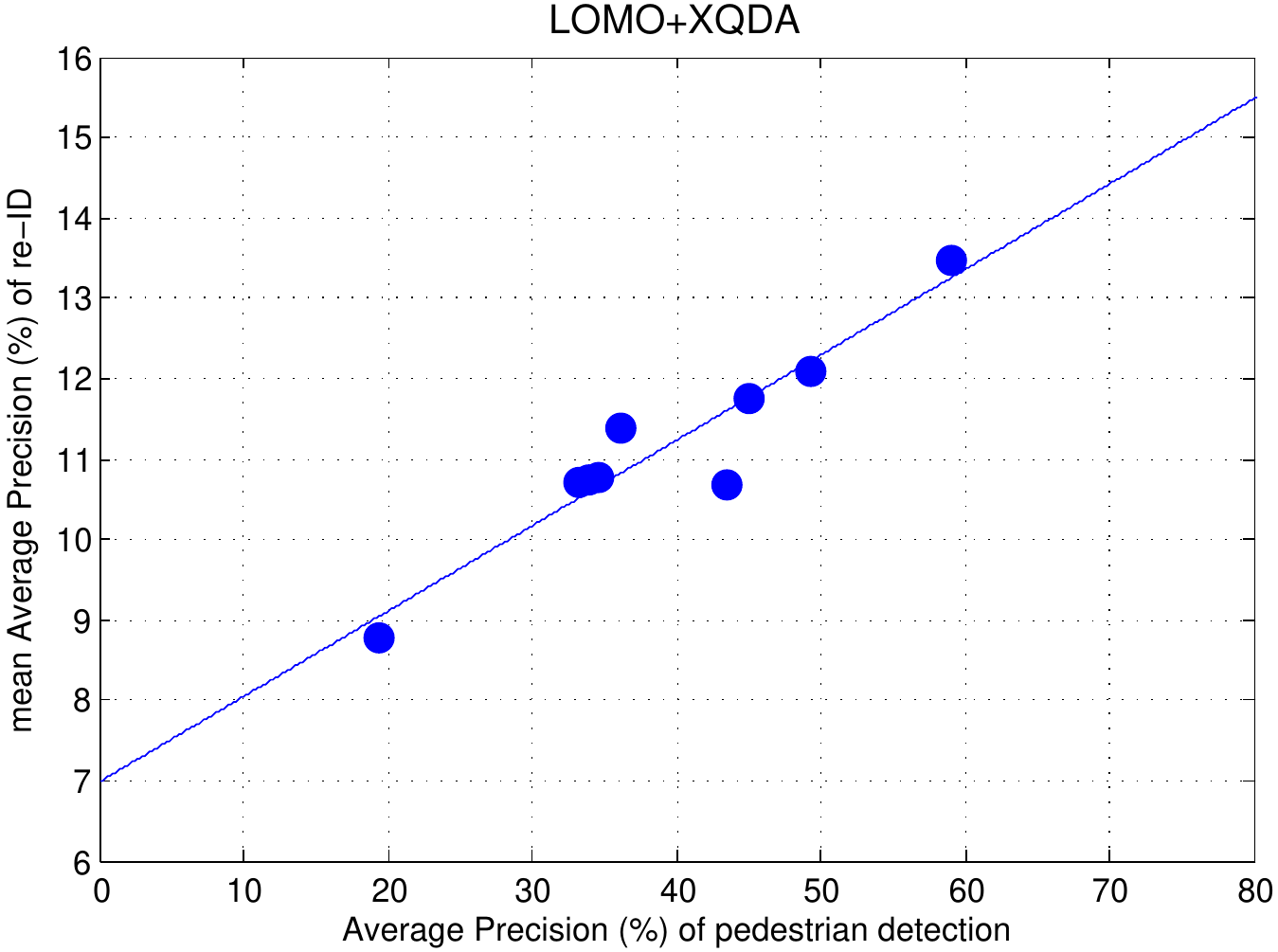}}

\subfigure[CNN, r1, IoU = 0.5]{\label{fig:viper}%
\includegraphics[width=1.7in]{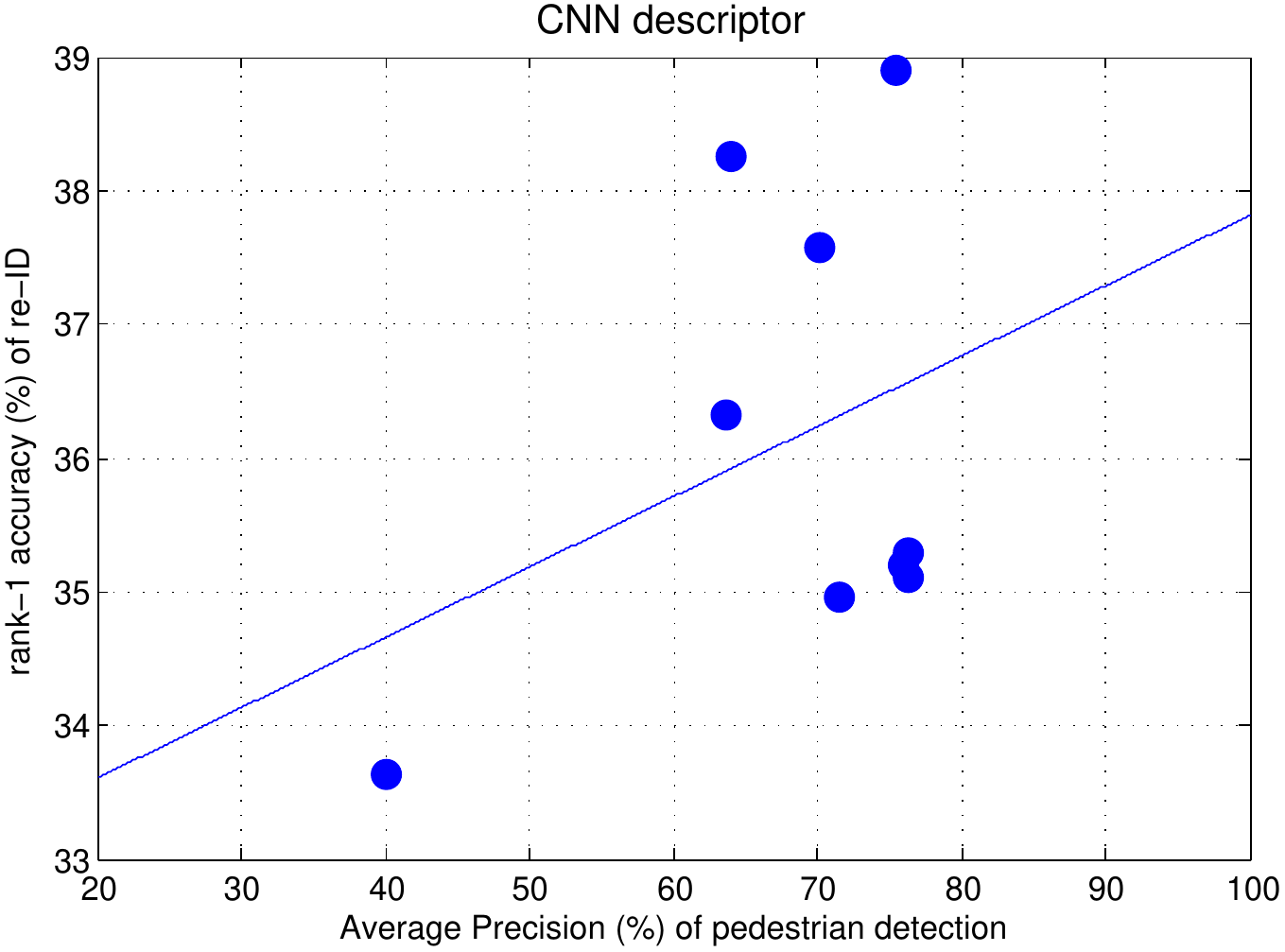}}
 \subfigure[CNN, r1, IoU = 0.7]{\label{fig:cuhk01}%
\includegraphics[width=1.7in]{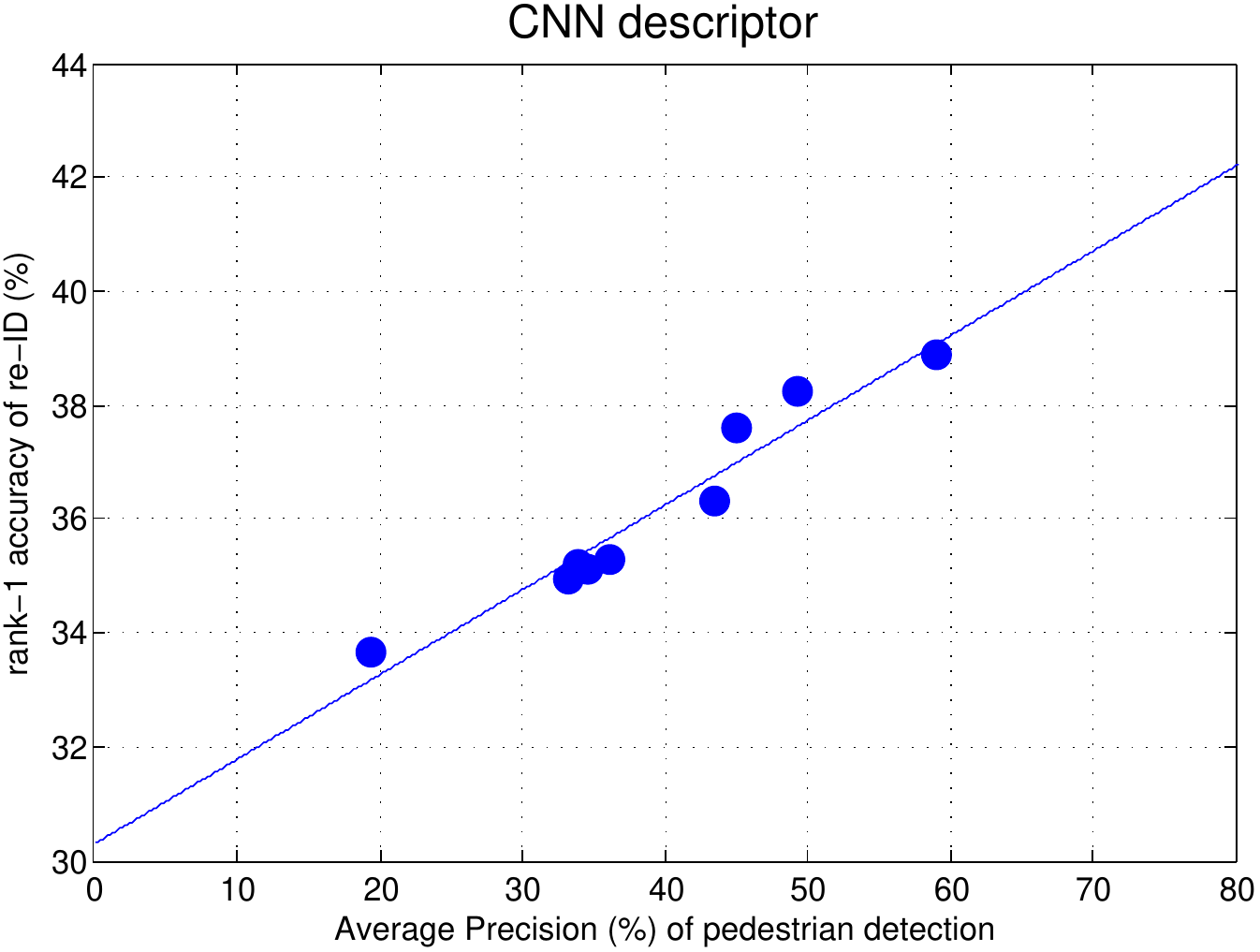}}
 \subfigure[CNN, mAP, IoU = 0.5]{\label{fig:iLIDS}%
\includegraphics[width=1.7in]{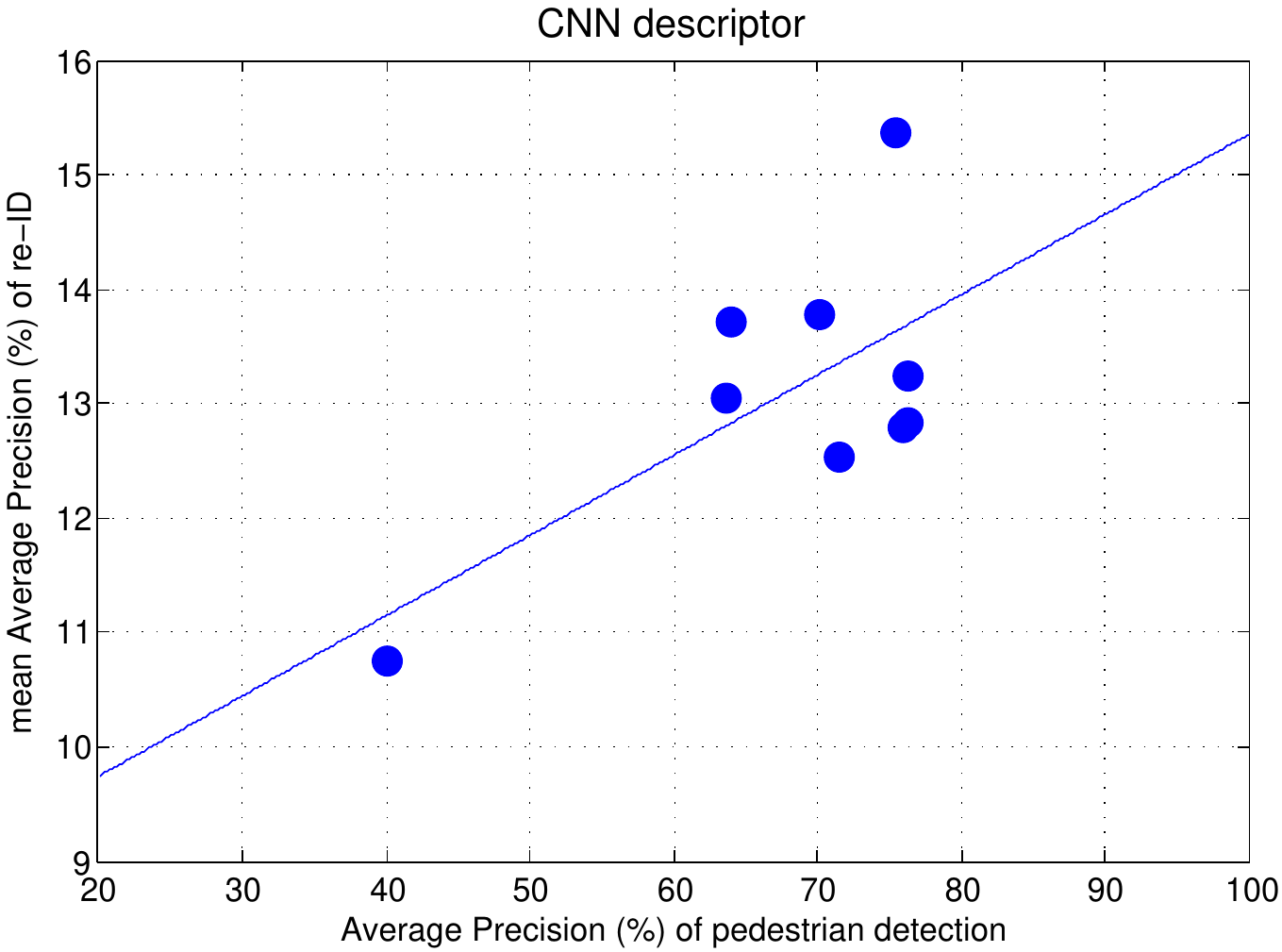}}
\subfigure[CNN, mAP, IoU = 0.7]{\label{fig:prid450s}%
\includegraphics[width=1.7in]{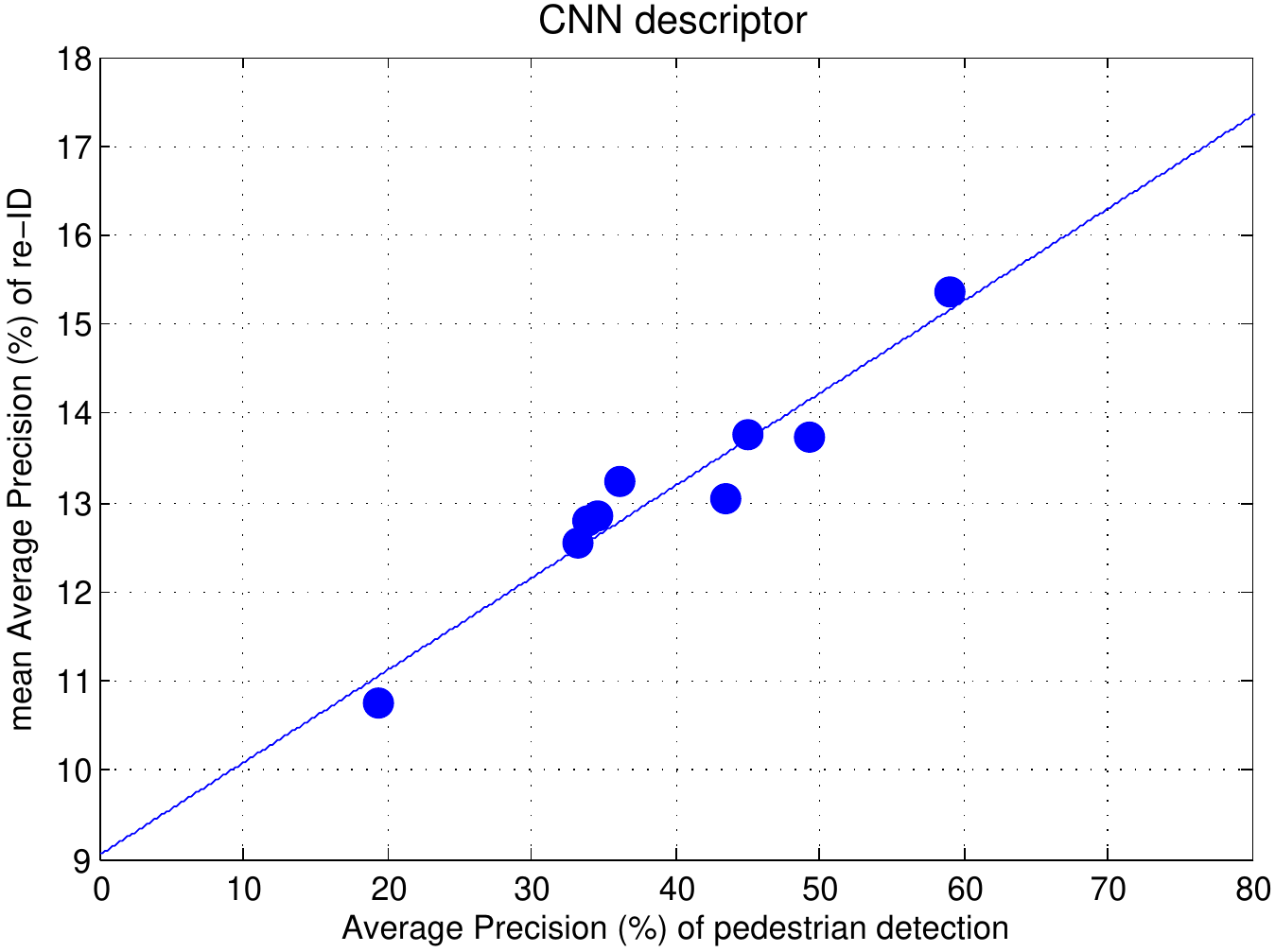}}
\caption{Person re-ID accuracy (mAP and rank-1) versus pedestrian detection accuracy (AP) on the PRW dataset \cite{zheng2016person}. Three re-ID methods are evaluated, \emph{i.e.,} BoW \cite{zheng2015scalable}, LOMO + XQDA \cite{liao2015person}, and CNN \cite{zheng2016person}. 9 detectors are evaluated, \emph{i.e.,} 1) DPM \cite{felzenszwalb2010object} + RCNN (AlexNet), 2) DPM pre-trained on INRIA \cite{dalal2005histograms}, 3) DPM re-trained on PRW, 4) ACF \cite{dollar2014fast} pre-trained on INRIA, 5) ACF + RCNN (AlexNet), 6) ACF + RCNN (ResidualNet), 7) ACF re-trained on PRW, 8) LDCF \cite{nam2014local} re-trained on PRW, and 9) LDCF pre-trained on INRIA. We can observe clearly the linear relation between re-ID and detection accuracy under IoU = 0.7 instead of IoU = 0.5.}
\label{fig:iou}
\end {figure*}
\subsection{Future Issues}

\subsubsection{System Performance Evaluation}
A proper evaluation methodology is a critical and sometimes tricky topic. Generally there is no single ``correct'' protocol, especially for the under-explored end-to-end re-ID task.
An end-to-end re-ID system departs from most current re-ID studies in  dynamic galleries based on the  specific detector/tracker used and their parameters. Moreover, it also  remains mostly unknown how to evaluate detection/tracking performance in the scenario of person re-ID. As a consequence, this survey raises questions of system evaluation on two aspects.

\setlength{\tabcolsep}{2.75pt}
\begin{table}[t]
\centering
\small
\caption{Datasets \cite{xiao2016end,zheng2016person,xu2014person,berclaz2011multiple} for end-to-end person re-identification (search).}
\begin{tabular}{l|cccc}
\hline

Dataset &LSPS&PRW&CAMPUS&EPFL\\
\hline
\#frames & 18,184 & 11,816& 214&80\\
\#ID &8,432 &932&74&30\\
\#annotated bbox & 99,809&34,304&1,519&294\\
\#box per ID &11.8 &36.8&20.5&9.8\\
\#gallery box &50-200k&50-200k&1,519&294\\
\#camera &-&6&3&4\\
Evaluation &CMC\&mAP&CMC\&mAP&CMC&CMC\\
\hline
\end{tabular}
\label{table:compare_end2end_dataset}
\end{table}
First, it is critical to use effective evaluation metrics for pedestrian detection and tracking in re-ID. The evaluation protocol should be able to quantify and rank detector/tracker performance in a realistic and unbiased manner and informative of re-ID accuracy. Pedestrian detection, for example, mostly employs the log-average miss rate (MR) which is averaged over the precision range of $[10^{-2}, 10^0]$ FPPI (false positives per image). Some also use average precision (AP) following the routine in PASCAL VOC \cite{everingham2010pascal}. Doll{\'a}r \emph{et al.} \cite{dollar2012pedestrian} argue that using the miss rate against FPPI is preferred to precision recall curves in certain tasks such as automotive applications, since there may be an upper limit on the acceptable FPPI. As opposed to the automotive applications of pedestrian detection, person re-ID aims to find a person which does not necessarily care about the false positive rates. So essentially we can employ both the miss rate and average precision to evaluate pedestrian detection for person re-ID.

An important parameter in the AP/MR computation is the intersect over union (IoU) score. A detected bounding box is considered correct if its IoU score with the ground truth bounding box is larger than a threshold. Typically the threshold is set to 0.5, and yet Zhang \emph{et al.} \cite{zhang2016far} study the difference between a ``perfect single frame detector'' and an automatic detector under various IoU scores. The KITTI benchmark \cite{geiger2012we} requires an IoU of 0.7 for car detection, but 0.5 for pedestrians. For person re-identification, this problem is open to proposals. Some clues about it still exist and if we dive closer to the conclusions drawn in \cite{zheng2016person}, we should be aware of the observation that using a larger IoU score (\emph{e.g.,} 0.7) is a better evaluation criteria than a low IoU (\emph{e.g.,} 0.5). Figure \ref{fig:iou} presents the relationship between detection accuracy (AP) and re-ID accuracy (rank-1 or mAP) on the PRW dataset. A linear relation is clearly presented between the two tasks under IoU = 0.7, while a scattered plot exists under IoU = 0.5. The correlation between detectors and recognizors is therefore more consistent with a larger IoU. Nevertheless, it is still far from satisfactory.

Given the consideration that bounding box localization quality is important for re-ID accuracy, it is a good idea to study IoU thresholds when assessing detector quality and see if it accords with re-ID accuracy. Our intuition is that a larger IoU criteria enforces better localization results, but there has to be some limit, because the difference in detector performance tends to diminish when IoU gets larger \cite{zhang2016far}. It would also feasible to explore the usage of the average recall (AR) proposed in \cite{hosang2016makes} for IoU from 0.5 to 1 and plot the AR for a varying number of detector thresholds. Such an evaluation metric considers both recall and localization, and we speculate that it may be especially informative in re-ID where pedestrian detection recall and bounding box quality are of vital importance.

While there are at least some clues to guide the evaluation of pedestrian detection, how to evaluate tracking under person re-ID is largely unknown. In the multiple object tracking (MOT) benchmark \cite{leal2015motchallenge}, multiple evaluation metrics are used, including multiple object tracking precision (MOTP) \cite{bernardin2008evaluating}, mostly track (MT) targets (percentage of ground truth persons whose trajectories are covered by the tracking results for at least 80\%), the total number of false positives (FP), the total number of ID switches (IDS), the total number of times a trajectory is fragmented (Frag), the number of frames processed per second (Hz), \emph{etc}. It might be possible that some of the metrics are of limited indication ability such as the processing speed, because tracking is an off-line step. For re-ID, we envision that tracking precision is critical as it is undesirable to have outlier images in the tracklets which compromise the effectiveness of pooling. We also speculate that 80\% might not be an optimal threshold for evaluating MT under re-ID. As suggested by \cite{wang2014person}, extracting features within a walking cycle is a good practise, so generating long tracking sequences may not bring much re-ID improvement. In the future, once datasets are released to evaluate tracking and re-ID, an urgent problem is thus to design proper metrics to evaluate different trackers.

The second question \emph{w.r.t} the evaluation procedure concerns the re-ID accuracy of the entire system. In contrast to traditional re-ID in which the gallery is fixed, in an end-to-end re-ID system, the gallery varies with the detection/tracking threshold. A stricter threshold indicates higher detection/tracking confidence, so the gallery is smaller and background detections are fewer and vice versa. Furthermore, the gallery size has a direct impact on re-ID accuracy. Let us take an extreme case as an example. When the detection/tracking threshold is very strict, the gallery can be very small, and it is even possible that the ground truth matches are excluded. At the other extreme, when the detection/tracking threshold is set to a very loose value, the gallery would be very large and contain a number of background detections which may exert a negative effect on re-ID, as shown in \cite{zheng2015scalable}. Therefore, it is predictable that too strict or too loose a threshold leads to inferior galleries, and it is preferred that the re-ID evaluation protocol reflect how the re-ID accuracy changes with the gallery dynamics. In \cite{zheng2016person}, Zheng \emph{et al.} plot rank-1 accuracy and mAP against a different number of detections per image. It is observed that the curves first rise and then drop after they peak. In the PRW dataset, the peak is positioned at 4-5 detections per images, which can serve as an estimation of the average number of pedestrians per image. In \cite{xiao2016end}, a similar protocol is employed, \emph{i.e.,} the rank-1 matching rate is plotted against detection recall, and reaches its maximum value when recall = 70\%. When recall further increases, the prevalence of false detections will compromise the re-ID accuracy. Some other ideas could be explored, \emph{e.g.,} plotting re-ID accuracy against FPPI. Keeping in mind that the gallery size depends on the detector threshold, other new evaluation metrics that are informative and unbiased could be designed in the future.

\begin{figure}[t]
  \centering
  \includegraphics[width=3.2in]{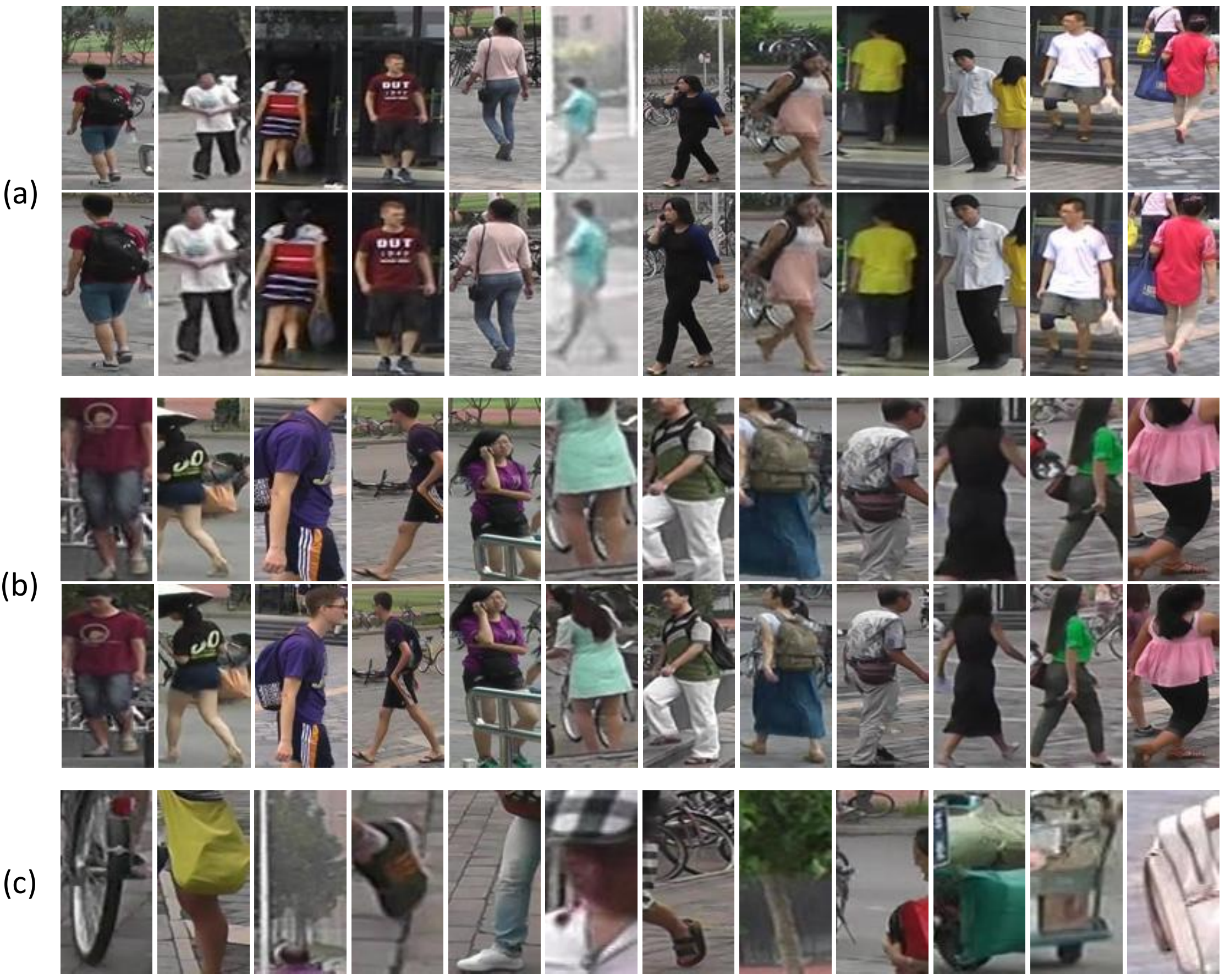}\\
  \caption{Detection errors in the Market-1501 dataset \cite{zheng2015scalable}. (a) misalignment and scale variances; (b) part missing; (c) false positives. In (a) and (b), the first and second rows represent DPM-detected and hand-drawn bounding boxes which have an IoU $>$ 0.5. }\label{fig:det_error}
\end{figure}
We also point out another re-ID evaluation protocol in end-to-end systems. In practice, when being presented with a query bounding box/video sequence, while it is good to locate the identity in a certain frame and tell its coordinates by pedestrian detection/tracking, it is also acceptable that the system only knows which frame(s) the identity re-appears in. The specific location of the query person can then be found by human labor which is efficient. In essence, determining the exact frame(s) where the queried person appears is a relatively easier task than a ``detection/tracking+re-identification'' task, because detection/tracking errors may not exert a large influence. In this scenario, re-ID accuracy should be higher than the standard re-ID task. Also, mean average precision can be used \emph{w.r.t} the retrieved video frames. Since this task does not require locating persons very precisely, we can thus use relaxed detectors/proposals or trackers aiming at improving frame-level recall. Detectors/proposals can be learned to locate a rough region of pedestrians with a loose IoU restriction, and put more emphasis on matching, \emph{i.e.,} finding a particular person from a larger bounding box/spatial-temporal tube.

\subsubsection{The Influence of Detector/Tracker on Re-ID} \label{sec:influence}
Person re-ID originates from pedestrian tracking \cite{huang1997object}, in which tracklets from multiple cameras are associated if they are determined to be of the same identity. This line of research treats re-ID as a part of the tracking system, and does not evaluate the impact of  localization/tracking accuracy on re-ID accuracy. However, even since the independence of re-ID, most studies have been conducted on hand-drawn image bounding boxes which is an idealized situation that hardly meets reality. Therefore, in an end-to-end re-ID system, it is critical that the impact of detection/tracking on re-ID be understood and that methods be proposed that detection/tracking methods/data can help re-ID.

First, pedestrian/tracking errors do affect re-ID accuracy, but the intrinsic mechanism and feasible solutions are still open to challenge. Detection errors (Fig. \ref{fig:det_error}) may lead to pedestrian misalignment, scale changes, part missing and most importantly, false positives and miss detections, which compromise the re-ID performance and pose new challenges for the community \cite{li2014deepreid,zheng2015scalable,wang2016human}.

A few re-ID works explicitly take  the detection/tracking errors into account. In \cite{zheng2015partial}, Zheng \emph{et al.} propose fusing local-local and global-local matches to address partial re-ID problems with severe occlusions or missing parts. In \cite{xu2014person}, Xu \emph{et al.} compute a ``commonness'' score by matching the GMM encoded descriptor with a prior distribution. The score can be used to eliminate false positives which do not contain or provide good localization of a human body. In a similar way, Zheng \emph{et al.} \cite{zheng2016person} propose integrating detector confidence (after square root) into the re-ID similarity score, according to which the gallery bounding boxes are ranked. These works address detection errors after they happen. Nevertheless, there is a possibility that detection/tracking errors could be avoided at an earlier stage. For example, in the network designed by Xiao \emph{et al.} \cite{xiao2016end}, a localization loss is added in the fast R-CNN \cite{girshick2015fast} sub-module. It regulates localization quality which is critical for an effective re-ID system.

Future investigations are in need to reveal the dependence of person re-ID on detection/tracking quality. Since the  idea to develop detector/trackers that are error-free is too idealistic, we advocate research into how detection confidence can be integrated into re-ID matching scores, \emph{i.e.,} how to correct errors by effectively identifying outliers, and how to train context models that do not rely solely on detected bounding boxes. For example, using clustering algorithms to filter out inconsistent frames within a tracklet can be effective in purifying tracking sequences. In another example, detected bounding boxes could be enlarged to include possibly missing body parts and learn discriminative visual features that in turn use or discard the enriched contextual information.

Secondly, we should be aware that detection and tracking, if appropriately designed, may be of help to re-ID. In \cite{zheng2016person}, the IDE network fine-tuned on the R-CNN model \cite{girshick2013rich} is proved to be more effective than the one fine-tuned directly on an ImageNet pre-trained model. This illustrates the importance of using the excessive amount of labeled data in pedestrian detection, \emph{i.e.,} pedestrians with ID annotations and false positive detections. In \cite{xiao2016end}, the end-to-end network integrates the loss of background detections, which is assumed to improve the discriminative ability of the learned embedding. The integration of detection scores into re-ID similarities \cite{xu2014person,zheng2016person} can also be viewed as an alternative that detection helps re-ID.

It may seem not quite straightforward that pedestrian detection/tracking could help re-ID or the reverse, but if we consider the analogy of generic image classification and fine-grained classification, we may think of some clues. From example, fine-tuning the ImageNet pre-trained CNN model on the fine-grained datasets is an effective way for faster convergence and higher fine-grained recognition accuracy. It is also a good idea to jointly train a pedestrian detection and re-ID model by back-propagating the re-ID loss to the (fast) RCNN part. Being able to discriminate different identities may be beneficial to the task of discriminating pedestrians from the background. The latter could also be helpful to the former.

One of the ideas that can be explored is the use of unsupervised tracking data. In videos, tracking a pedestrian is not too difficult a task, though tracking errors are inevitable. Facial recognition, color, and non-background information are useful tools to improve tracking performance like in Harry Potter's Marauder's Map \cite{yu2013harry}.  Within a tracking sequence, the appearance of a person may undergo variances to some extent, but it can be expected that most of the bounding boxes are of the same person. In this scenario, each tracklet represents a person which contains a number of noisy but roughly usable training samples. We can therefore make use of racking results to train pedestrian verification/identification models, so as to alleviate the reliance on large-scale supervised data. As another promising idea, it is worth trying to pre-train CNN models using the detection/tracking data using the auto-encoder \cite{deng2010binary} or the generative adversarial nets (GAN) \cite{goodfellow2014generative}. It would also be interesting to directly learn person descriptors using such unsupervised networks to help address the data issue in person re-ID.

\section{Future: Person Re-ID in Very Large Galleries}\label{sec:scale_reid}
The scale of data has increased significantly in the re-ID community in recent years, \emph{e.g.,} from several hundred gallery images in VIPeR \cite{gray2007evaluating} and iLIDS \cite{zheng2009associating} to over 100k as in PRW \cite{zheng2016person} and LSPS \cite{xiao2016end}, which gives rise to the predominance of deep learning methods. However, it is apparent that current datasets are still far from a practical scale. Supposing that in a region-scale surveillance network with 100 cameras, if one video frame is used per second for pedestrian detection, and an average of 10 bounding boxes are produced from each frame, then, running the system for 12 hours will produce $3,600\times 12 \times 1 \times 10 \times 100 = 43.2\times10^6$ bounding boxes. But to our knowledge, previously no work has reported re-ID performance in such a large gallery. It seems that the largest gallery used in the literature is 500k \cite{zheng2015scalable}, and evidence suggests that mAP drops over 7\% compared to Market-1501 with a 19k gallery. Moreover, in \cite{zheng2015scalable}, approximate nearest neighbor search \cite{wang2012query} is employed for fast retrieval but at the cost of compromised accuracy.

From both a research and an application perspective, person re-ID in very large galleries should be a critical direction in the future. Attempts to improve both the accuracy and efficiency issues should be made.

\begin{figure}[t]
  \centering
  \includegraphics[width=3.5in]{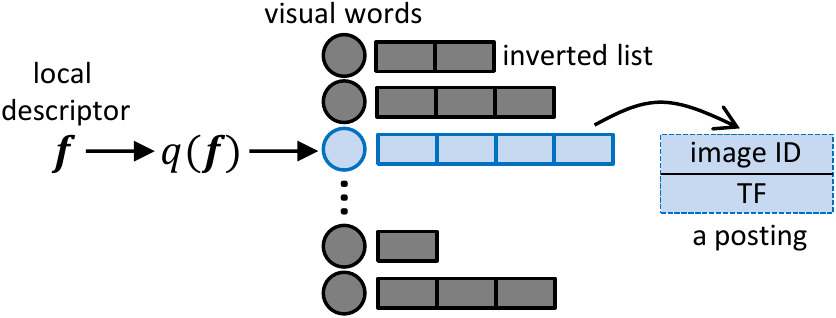}\\
  \caption{An illustration of the inverted index.}\label{fig:inv_index}
\end{figure}
On the one hand, robust and large-scale learning of descriptors and distance metrics is much more important. This coincides with current research \cite{varior2016siamese,liu2016end,su2016deep,wu2016enhanced}. Following large-scale image recognition \cite{deng2009imagenet}, person re-ID will progress to large-scale evaluations. Although current methods address the re-ID problem between one or several pairs of cameras in a very limited time window, robustness in a camera networks over a long time period has not been well considered. In \cite{das2014consistent,chakraborty2014network}, the re-ID consistency within a camera network is jointly optimized with pair-wise matching accuracy, but the testing datasets (WARD \cite{martinel2012re} and RAiD \cite{das2014consistent} ) only have 3 and 4 cameras and less than 100 identities. In a network with $n$ cameras, the number of camera pairs is $\mathcal{O}(n^2)$. Considering the long recording time and lack of annotated data, it is typically prohibitive to train distance metrics or CNN descriptors in a pair-wise manner. As a consequence, training a global re-ID model with adaptation to various illumination condition and camera location is a priority. Toward this goal, an option is to design unsupervised descriptors \cite{ma2012bicov,zheng2015scalable} which aim to find visually similar persons and treat visually dissimilar ones as false matches. But unsupervised methods may be prone to lighting changes.

On the other hand, efficiency is another important issue in such a large-scale setting. Although computation time could almost be omitted in small datasets \cite{gray2007evaluating,zheng2009associating}, in our experiment using MATLAB 2014 on a server with 3.1GHz Intel Xeon E5-2687w v3 (10 cores), 64GB memory, it takes 8.50s to compute the distance between a 100-dim floating vector with a number of 10 million 100-dim vectors. If we use a 4,096-dim floating-point vector extracted from the CaffeNet \cite{krizhevsky2012imagenet} and C++ programming, the time used increases dramatically to 60.7s including 33.2s for the distance calculation and 26.8s for the data to load  from the disk. It is clear that the query time increases dramatically according to the feature dimensions and gallery size, which is not desirable for practical use. To our knowledge, previous works in person re-ID rarely focus on efficiency issues, and therefore effective solutions are lacking, but fortunately, we can resort to the image retrieval community for answers, and this survey provides two possible directions.

\textbf{Inverted index-based.} The inverted index is a \emph{de facto} data structure in the Bag-of-Words (BoW) based retrieval methods \cite{AKM,HKM,Hamming}. Based on the quantization result of local descriptors, the inverted index has $k$ entries where $k$ denotes the codebook size. The indexing structure thus has $k$ entries and each is attached to an inverted list, in which the local descriptors are indexed. The structure of the baseline inverted index is shown in Fig. \ref{fig:inv_index}. A posting stores the image ID and the term frequency (TF) of the indexed descriptor and in a series of works, a number of other meta data can be stored, such as binary signature \cite{Hamming}, feature coordinates \cite{GVP}, \emph{etc}.  For basic knowledge and state-of-the-art advances of the inverted index in instance retrieval, we refer readers to a recent survey \cite{zheng2016sift}.

In person re-ID, the use of local descriptors is popular \cite{zheng2015scalable,zhao2013unsupervised,zhao2014learning}. The color and texture features are typically extracted from local patches. While some previous works use sophisticated matching algorithms \cite{zhao2013unsupervised}, it is preferred that the matching procedure can be accelerated using the inverted index under a large gallery. A codebook is usually needed to quantize a local descriptor to visual words, and since the local descriptors are high-dimensional, a large codebook is needed to reduce quantization error. Under these circumstances, the inverted index is ready for use which saves memory costs to a large extent and, if properly employed, can have approximately the same accuracy compared to quantization-free cases.

\textbf{Hashing-based.} Hashing has been an extensively studied solution to approximate nearest neighbor search, which aims to reduce the cost of finding exact nearest neighbors when the gallery is large or distance computation is costly \cite{wang2016survey}. Learning to hash is popular in the community following the milestone work Spectral Hashing  \cite{weiss2009spectral}. It refers to learning hash functions, $y = h(x)$, mapping a vector $x$ to a compact $y$, and aims at finding the true nearest neighbor at high-ranks in the rank list while keeping the efficiency of the search process. Some classic learning to hash methods include product quantization (PQ) \cite{jegou2010aggregating}, iterative quantization (ITQ) \cite{gong2011iterative}, \emph{etc}. Both methods are efficient in training and have fair retrieval accuracy. They do not require labeled data, so are applicable for re-ID tasks when large amount of training data may not be available.

Another application of supervised hashing consists of image retrieval \cite{liu2016multilinear,zhang2016efficient,zhao2015deep,erin2015deep}, which is the interest of this section. The hash function is learned end-to-end through a deep learning network which outputs a binary vector given an input image. This line of works focus on several image classification datasets such as CIFAR-10 \cite{krizhevsky2009learning} and NUS-WIDE \cite{chua2009nus}, in order to leverage the training data that is lacking in generic instance retrieval datasets \cite{Hamming,AKM}. In person re-ID, the application scenario fits well with deep hashing for image retrieval. In large galleries, efficient yet accurate hash methods are greatly needed, which is a less-explored direction in re-ID. As shown in Table \ref{table:compare_task}, training classes are available in re-ID datasets, and the testing procedure is a standard retrieval task, so the current arts in supervised hashing are readily to be adopted in re-ID in the light of the increasing size of the datasets \cite{zheng2015scalable,li2014deepreid}. The only relevant work we find is \cite{zhang2015bit} which learns hash functions in a triplet-loss CNN network with regularizations to enforce adjacency consistency. This method is tested on the CUHK03 dataset which contains 100 identities in each test split, so in this sense, performance evaluation on very large galleries is still lacking. As a consequence, this survey calls for very large re-ID datasets that will evaluate the scalability of re-ID methods and scalable algorithms especially those using hash codes to further push this task to real-world applications.

\section{Other Important Yet Under-developed Open Issues}\label{sec:open}
\subsection{Battle Against Data Volumn} \label{sec:data_volumn}
Annotating large-scale datasets has always been a focus in the vision community. This problem is even more challenging in person re-ID, because apart from drawing a bounding box of a pedestrian, one has to assign him an ID. ID assignment is not trivial since a pedestrian may re-enter the fields of view (FOV) or enter another observation camera a long time after the pedestrian's first appearance. This makes collaborative annotation difficult, as it is costly for two collaborators to communicate on the annotated IDs. These difficulties partially explain why current datasets typically have a very limited number of images for each ID. The last two years have witnessed the release of several large-scale datasets, \emph{e.g.,} Market-1501 \cite{zheng2015scalable}, PRW \cite{zheng2016person}, LSPS \cite{xiao2016end}, and MARS \cite{zheng2016mars}, but they are still far from satisfaction in views of practical applications. In this survey, we believe two alternative strategies can help bypass the data issue.

First, how to use annotations from tracking and detection datasets remains under-explored. Compared to re-ID, tracking and detection annotations do not require ID assignment when a person re-enters FOV: the majority of effort has been spent on bounding box drawing. In \cite{zheng2016person}, it is shown that adding more pedestrian and background training data in the R-CNN stage benefits the following training of the IDE descriptor. In \cite{su2016deep,su2015multi}, attribute annotations from independent datasets are employed to represent the re-ID images. Since the attributes can be annotated through collaboration among workers and have good generalization ability, they are also good alternatives to the deficiency of re-ID data. As a consequence, external resources are valuable for training re-ID systems when training data is lacking.

\begin{figure}[t]
  \centering
  \includegraphics[width=3.5in]{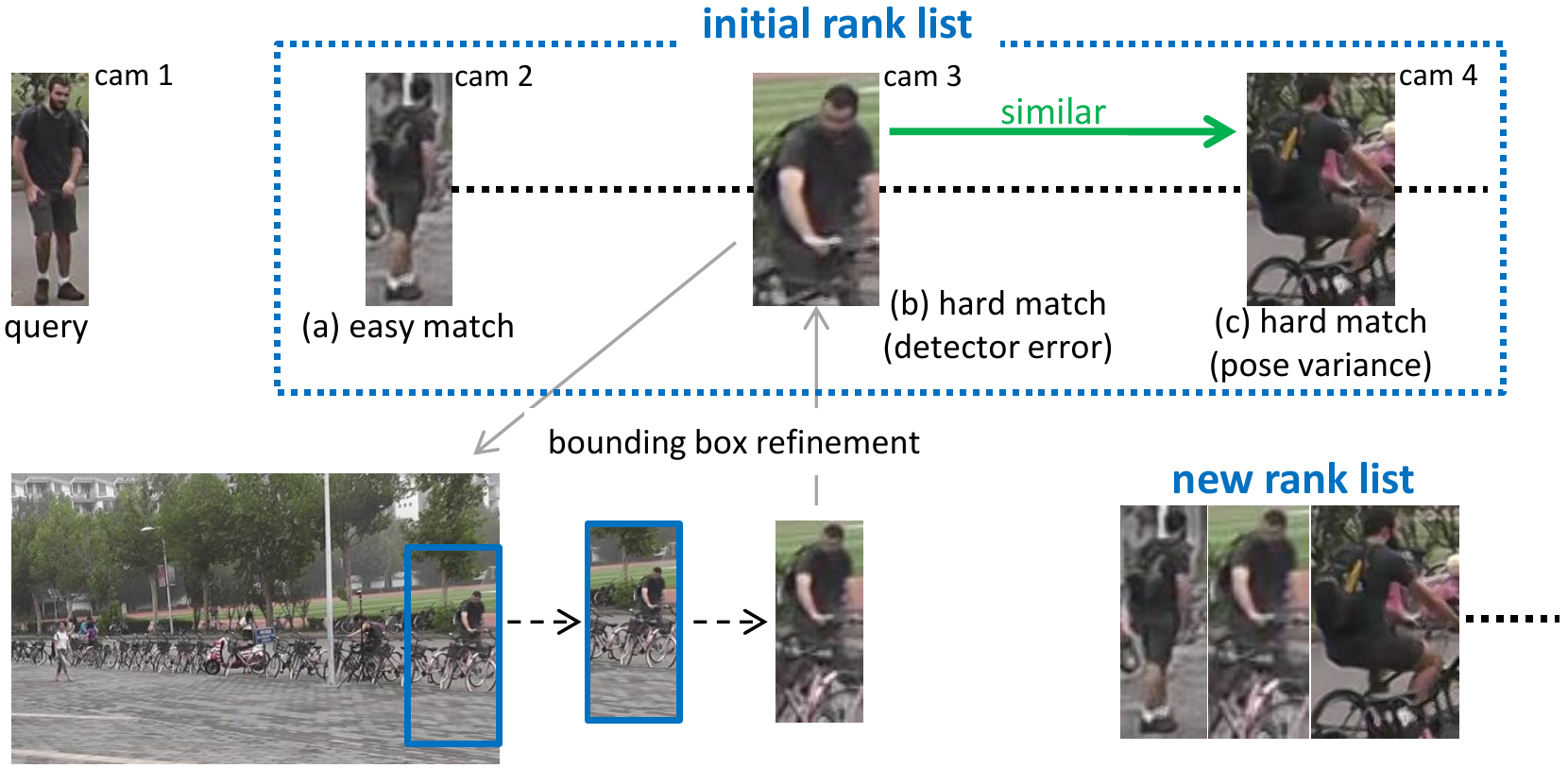}\\
  \caption{An example of re-ranking in re-ID. Given a query image, an initial rank list is obtained, in which an easy match (a) is ranked top, while two hard matches (b) and (c) have low ranks. The detection error in (b) can be corrected by retrieving the corresponding video frame and performing a finer search for the best bounding box within a local neighborhood. In this example, (c) is visually similar to (b) but not the query, so after (b) is retrieved, (c) can be found by similarity propagation.}\label{fig:reranking}
\end{figure}

Apart from the pre-training/unsupervised strategies as mentioned in Section \ref{sec:influence}, a novel solution is to retrieve hard negatives from the unlabeled data which can be viewed as ``true positives'' in metric learning/CNN training. This strategy has been evaluated in object classification where a small portion of labels are disturbed before training \cite{xie2016disturblabel}. It can efficiently enlarge the training set, and at the same time reduce the risk of model over-fitting. Our preliminary experiments show that this direction yields decent improvement over the baselines.

The second strategy is transfer learning that transfers a trained model from the source to the target domain. Previously, supervised learning require large numbers of labeled data which limits the re-ID system to scale to other cameras. In \cite{wang2014unsupervised}, an unsupervised topic model is proposed to discover saliant image patches for re-ID matching and simultaneously remove background clusters. In \cite{wang2015cross}, a weakly supervised method is proposed which requires full annotations from other re-ID dataset and a few samples captured in the target scenario. In \cite{ma2015cross,peng2016unsupervised}, unsupervised transfer learning is proposed in which the target dataset is unlabeled. Ma \emph{et al.} \cite{ma2015cross} employ a cross-domain ranking SVM, while Peng \emph{et al.} \cite{peng2016unsupervised} formulate the transfer problem as a dictionary learning task, which learns the shared invariant latent variables and is biased towards the target dataset. These methods indicate that it is feasible to learn a fair re-ID model from the source, and that it is beneficial to mine discriminative cues from the unsupervised data. Transfering CNN models to other re-ID datasets can be more difficult because the deep model provides a good fit to the source. Xiao \emph{et al.} \cite{xiao2016learning} gather a number of source re-ID datasets and jointly train a recognition model for the target dataset. According to our experience, the usage of off-the-shelf metric learning methods \cite{liao2015person,kostinger2012large} can also correct the transfer effect to some extent, but unsupervised transfer learning is still an open issue for the deeply learned models.

\subsection{Re-ranking Re-ID Results} \label{sec:re-ranking}
The re-identification process (Fig. \ref{fig:pipeline}(b)) can be viewed as a retrieval task, in which re-ranking is an important step to improve the retrieval accuracy. It refers to the re-ordering of the initial ranking result from which re-ranking knowledge can be discovered. For a detailed survey of search re-ranking methods, we refer the readers to \cite{mei2014multimedia}.

A few previous works exist on this topic. Re-ranking can be performed either with human in the loop or fully automatically. When online human labeling is involved, Liu \emph{et al.} \cite{liu2013pop} propose the post-rank optimisation (POP) method which allows a user to provide an easy negative and, optionally, a few hard negatives from the initial rank list. The sparse human feedback enables on-the-fly automatic discriminative feature selection of the query person. In an improvement, Wang \emph{et al.} \cite{wang2016human} design the human verification incremental learning (HVIL) model which does not require any pre-labelled training data and learns cumulatively from human feedback to provide instance model update. A number of incrementally learned HVIL models are combined into a single ensemble model for use when human feedback is no longer available. In a similary nature, Martinel \emph{et al.} \cite{martinel2016temporal} propose finding the most relevant gallery images for a query, sending them to the human labeler, and finally using the labels to update the re-ID model. Automatic re-ranking methods have also been studied in several works. Zheng \emph{et al.} \cite{zheng2015query} propose a query-adaptive fusion method to combine rank results of several re-ID systems. Specifically, the shape of the initial score curves is used and it is argued that the curve exhibits an ``L'' shape for a good feature. In \cite{paisitkriangkrai2015learning}, various metrics are ensembled based on the direct optimization of the CMC curve.  Garc{\'i}a \emph{et al.} \cite{garcia2015person} analyze the unsupervised discriminant context information in the rank list. This is further combined with a re-ranking metric learned in the offline. Leng \emph{et al.} \cite{leng2015person} use the idea of reciprocal $k$-nearest neighbors \cite{qin2011hello} to refine the initial rank list based constructing images relations in the offline steps.

Re-ranking is still an open direction in person re-ID, while it has been extensively studied in instance retrieval. The application scenario can be depicted as follows. When searching for a person-of-interest, it is likely that its images under certain cameras are very difficult to find due to intensive image variations. But we may be able to find the true matches under some cameras which are more similar to the hard positives. So in this manner, hard positives can be found once the easy ones are returned. Re-ranking methods in instance retrieval can be readily adopted in person re-ID \cite{qin2011hello,zheng2015scalable,shen2012object,root_sift}. Since training data is available in re-ID (Table \ref{table:compare_task}), it is possible to design re-ranking methods based on training distribution. For example, when doing k-NN re-ranking \cite{shen2012object}, the validity of the returned results can be determined from the training set according to the scores. Since re-ID is focused on pedestrians, re-ranking methods can be specifically designed. For example, after obtaining the initial rank list, a subset of the top-ranked images can be selected, and the video frames containing them can be retrieved. We can subsequently find the best localization through expensive sliding window method without incurring much computation burdens, so as to allieviate the impact of detector misalignment.

\subsection{Open-World Person Re-ID} \label{sec:open-world}
Most existing re-ID works can be viewed as an identification task (Eq. \ref{eq:img_reid}). Query identities are assumed to exist in the gallery and the tasks aim to determine the ID of the query. By contrast, open-world re-ID systems study the person verification problem.
That is, based on identification tasks, the open-world problem adds another condition to Eq. \ref{eq:img_reid},
\begin{equation}\label{eq:open-world}
  \mbox{sim}(q,g_{i^*}) > h,
\end{equation}
where $h$ is the threshold above which we can assert that query $q$ belongs to identity $i^*$; otherwise, $q$ is determined an outlier identity which is not contained in the gallery, although $i^*$ is the first ranked identity in the identification process.

In literature, open-world person re-ID is still at its early stage, and several works are proposed to help define this task. In \cite{zheng2016towards}, Zheng \emph{et al.} design a system consisting of a watch list (gallery) of several known identities and a number of probes including target and non-target ones. Their work aims to achieve high true target recognition (TTR) and low false target recognition (FTR) rate which calculate rate of the number of queries that are verified as the target identities to the total number of queries. In \cite{liao2014open}, Liao \emph{et al.} divide open-world re-ID into two sub-tasks, \emph{i.e.,} detection and identification; the former decides whether a probe identity is present in the gallery and the latter assigns an ID to the accepted probe. Consequently two different evaluation metrics, the detection and identification rate (DIR) and the false accept rate (FAR) are proposed, based on which a receiver operating characteristic (ROC) curve can be drawn.

Open-world re-ID still remains a challenging task as evidenced by the low recognition rate under low false accept rate, as shown in \cite{zheng2016towards,liao2014open}. The challenge mainly lies in two aspects \emph{i.e.,} detection and recognition, both of which are limited to the unsatisfying matching accuracy - a research focus in standard re-ID tasks. As indicated in \cite{liao2014open}, a 100\% FAR corresponds to the standard close-set re-ID and its accuracy is limited by the current state of the art; a lower FAR is accompanied by lower re-ID accuracy due to the low recall of the true matches. As a consequence, from a technical perspective, the critical goal is to improve matching accuracy, based on which probabilistic models can be designed for novelty detection (verification) methods. Moreover, when focusing on re-ID accuracy, open-world re-ID should also consider the dynamics of the gallery \cite{decann2015modelling}. In a dynamic system with constantly incoming bounding boxes, a new identity will be added to the ``watch list'' if it is determined to not belong to any existing gallery identities, and vice versa. Enrolling new identities dynamically enables automatic database construction and facilitates the re-ID process with a  pre-organized gallery.

\section{Concluding Remarks}\label{sec:conclusion}
Person re-identification, foretold in the oldest stories, is gaining extensive interest in the modern scientific community.  In this paper, a survey of person re-identification is presented. First, a brief history of person re-ID is introduced and its similarities and differences to image classification and instance retrieval are described. Then, existing image and video-based methods are reviewed, which are categorized into hand-crafted and deeply-learned systems. Positioned inbetween image classification and instance retrieval, person re-ID has a long way from becoming an accurate and efficient application. Therefore, departing from previous surveys, this paper places more emphasis on the under-developed but critical future possibilities, such as the end-to-end re-ID systems that integrate pedestrian detection and tracking, and person re-ID in very large galleries, which we believe are necessary steps toward practical systems. We also highlight some important open issues that may attract further attention from the community. They include solving the data volume issue, re-ID re-ranking methods, and open re-ID systems. All in all, the integration of discriminative feature learning, detector/tracking optimization, and efficient data structures will lead to a successful person re-identification system.


\ifCLASSOPTIONcompsoc
  \section*{Acknowledgments}
\else
  \section*{Acknowledgment}
\fi

The authors would like to thank the pioneer researchers in  person re-identification and other related fields.

\ifCLASSOPTIONcaptionsoff
  \newpage
\fi

\bibliographystyle{IEEEtran}
\bibliography{egbib}

\end{document}